%% file: acl_latex.tex
\pdfoutput=1

\documentclass[11pt]{article}

\usepackage[final]{acl}

\usepackage{times}
\usepackage{latexsym}

\usepackage[T1]{fontenc}

\usepackage[utf8]{inputenc}

\usepackage{microtype}

\usepackage{inconsolata}

\usepackage{graphicx}

\usepackage{adjustbox}
\usepackage{booktabs}
\usepackage{amsmath}
\usepackage{amssymb}
\usepackage{pifont}
\usepackage{subcaption,caption}
\usepackage{multirow}
\usepackage{fontawesome}
\usepackage{bbding}
\usepackage{tcolorbox}

\definecolor{customred}{HTML}{b22222}
\definecolor{customgreen}{HTML}{5f9ea0}

\title{SConU: Selective Conformal Uncertainty in Large Language Models}

\author{
 \textbf{Zhiyuan Wang\textsuperscript{\ding{170},$\dagger$}},
 \textbf{Qingni Wang\textsuperscript{\ding{170},$\dagger$}},
 \textbf{Yue Zhang\textsuperscript{$\clubsuit$}},
 \textbf{Tianlong Chen\textsuperscript{$\spadesuit$}},
\\
 \textbf{Xiaofeng Zhu\textsuperscript{\ding{170}}},
 \textbf{Xiaoshuang Shi\textsuperscript{\ding{170},$\ast$}},
 \textbf{Kaidi Xu\textsuperscript{$\clubsuit$,}\thanks{Corresponding Authors \textsuperscript{$\dagger$}Equal Contribution}}
\\
\\
 \textsuperscript{\ding{170}}University of Electronic Science and Technology of China
\\
 \textsuperscript{$\clubsuit$}Drexel University
\\
 \textsuperscript{$\spadesuit$}University of North Carolina at Chapel Hill
\\
\small \texttt{\{yhzywang, qingni1031, seanzhuxf, xsshi2013\}@gmail.com} \\
    \small \texttt{\{yz899 ,kx46\}@drexel.edu}\quad \texttt{tianlong@cs.unc.edu}
}

\begin{document}
\maketitle
\begin{abstract}
As large language models are increasingly utilized in real-world applications, guarantees of task-specific metrics are essential for their reliable deployment. 
Previous studies have introduced various criteria of conformal uncertainty grounded in split conformal prediction, which offer user-specified correctness coverage. 
However, existing frameworks often fail to identify uncertainty data outliers that violate the exchangeability assumption, leading to unbounded miscoverage rates and unactionable prediction sets. 
In this paper, we propose a novel approach termed Selective Conformal Uncertainty (SConU), which, for the first time, implements significance tests, by developing two conformal p-values that are instrumental in determining whether a given sample deviates from the uncertainty distribution of the calibration set at a specific manageable risk level. 
Our approach not only facilitates rigorous management of miscoverage rates across both single-domain and interdisciplinary contexts, but also enhances the efficiency of predictions. 
Furthermore, we comprehensively analyze the components of the conformal procedures, aiming to approximate conditional coverage, particularly in high-stakes question-answering tasks.\footnote{The code implementation for our experiments is available at \href{https://github.com/Zhiyuan-GG/SConU}{https://github.com/Zhiyuan-GG/SConU}} 
\end{abstract}

\input{Sections/Introduction}
\input{Sections/RelatedWork}
\input{Sections/Method}

\input{Sections/Experiments}

\input{Sections/Conclusion}




\section*{Acknowledgments}
Zhiyuan Wang, Xiaofeng Zhu and Xiaoshuang Shi were supported by the National Key Research \& Development Program of China under Grant (No. 2022YFA1004100), and the National Natural Science Foundation of China (No.62276052).


\section*{Limitations}
Our SConU framework excludes test QA samples that significantly deviate from the uncertainty distribution of the calibration set. 
In future work, we will investigate strategies to address nonexchangeable data sequences by analyzing the degree of uncertainty distribution shift between the given test sample and the calibration set. Moreover, we achieve approximate conditional coverage at various prediction set sizes in high-stakes QA tasks, prompting us to conduct more comprehensive studies on the mechanisms influencing conditional performance on particular data points in subsequent research. 

\newpage

\bibliography{custom}

\newpage

\appendix

\input{Sections/Appendix}

\end{document}

%% file: Sections/Introduction.tex
\section{Introduction}
\label{sec: Introduction}
Large language models (LLMs) have been increasingly deployed in real-world natural language generation (NLG) tasks, including question-answering (QA)~\citep{duan-etal-2024-shifting,WANG2025109553}. 
However, their generations often reveal deficiencies in trustworthiness and robustness~\citep{yao2024survey,yona2024can,farquhar2024detecting,kaur2024addressing,pmlr-v235-hong24a}. 
These issues have sparked significant interest in developing guarantees for task-specific performance metrics, such as correctness miscoverage rate~\citep{wang-etal-2024-conu,quach2024conformal,wang2024sample}, factuality~\citep{pmlr-v235-mohri24a,cherian2024large}, and disparities in generation quality across diverse user populations~\citep{deng2023distributionfree,zollo2024prompt}.

Split conformal prediction (SCP)~\citep{splitconformal,bates2021distribution,angelopoulos2021gentle} offers distribution-free and model-agnostic coverage guarantees to new samples based on a calibration set. 
Recent studies have introduced various criteria of conformal uncertainty (ConU), which allow user-specified risk levels (e.g., $\alpha$) for the coverage of acceptable responses in practical NLG tasks, by correlating the nonconformity score (NS) with the uncertainty state of ground-truth answers~\citep{quach2024conformal,su-etal-2024-api,wang-etal-2024-conu,wang2024sample,kaur2024addressing}. 
However, these frameworks are vulnerable to uncertainty outliers and sensitive to internal units, such as the uncertainty notion and split ratio, compromising their statistical rigor and operational efficiency~\citep{cresswell2024conformal,plassier2024probabilistic}. 

\begin{figure*}[!t]
    \centering
    \begin{subfigure}{0.45\linewidth}
        \centering
        \includegraphics[width=\linewidth]{Figures/mmlu_pro.pdf}
        \caption{Single-domain Miscalibration.}
	\label{fig: nonexchangeable-single}
    \end{subfigure}
    \hfill
    \centering
    \begin{subfigure}{0.54\linewidth}
	\centering
	\includegraphics[width=\linewidth]{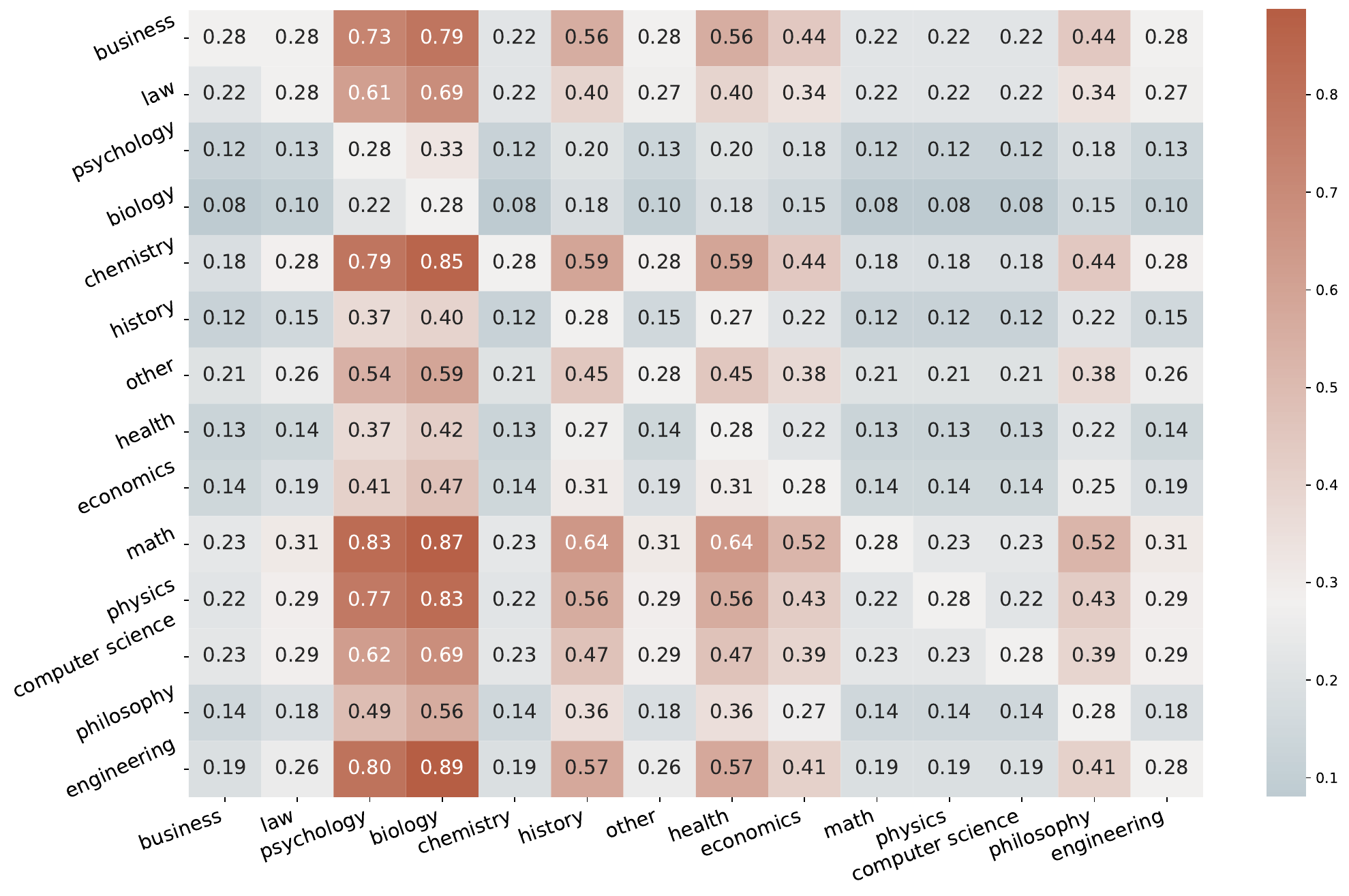}
	\caption{Cross-domain Miscalibration.}
	\label{fig: nonexchangeable-cross}
    \end{subfigure}
    \caption{\textbf{(a)} Empirical miscoverage rate (EMR) at various risk levels on the MMLU-Pro dataset utilizing 8 LLMs. Results on the left are from the Health discipline, while results on the right are from the Economics discipline. Solid lines give the mean over 100 trials and shaded regions show $+ / -$ the standard deviation (std). We set the split ratio between the calibration and test set to 0.5 for all trials. The {\color{customred} $\bigstar$} indicates that even the mean miscoverage rate at the corresponding risk level is higher than the upper bound, and the shaded regions exceeding the upper bound reflect significant data point anomalies. \textbf{(b)} Significant violation in the management of EMR when we use data points from different disciplines for the calibration set and the test set within the MMLU-Pro dataset employing the LLaMA-3.1-8B-Instruct model at the risk level of 0.28. Note that we calculate the minimum reliable risk level on each subject based on Eq.~\eqref{eq: minimum risk level} and set $\alpha$ to the maximum. All data on the diagonal is manually set to equal $\alpha$.}\label{fig: nonexchangeable-check}
\end{figure*}

To conduct comprehensive research,  we first revisit a crucial precondition for prior frameworks: the combined sequence of the given test QA sample and all calibration data points should be exchangeable~\citep{kumar2023conformal}. 
In practical QA tasks, however, this condition is hard to characterize and verify specifically, often being violated due to the conditional nature of language generation approaches~\citep{ulmer-etal-2024-non}. 
More concerning, we observe significant coverage anomalies within single-domain contexts, as illustrated in Figure~\ref{fig: nonexchangeable-single}, which contradict the assumptions made in previous studies~\cite{ye2024benchmarking,quach2024conformal,su-etal-2024-api,wang-etal-2024-conu,kaur2024addressing,wang2024sample}. 
Furthermore, miscalibration issues become even more pronounced in interdisciplinary scenarios~\citep{kumar2023conformal}, as demonstrated in Figure~\ref{fig: nonexchangeable-cross}. 
The conceptual and fragile nature of exchangeability renders the prediction sets produced by existing ConU frameworks unreliable and less actionable~\citep{cresswell2025conformal}.

Within prior ConU frameworks, the NSs are derived from various uncertainty notions linked with reliable generations and then utilized to select responses by a user-specified quantile. 
As supported by Figure~\ref{fig: nonexchangeable-check}, our key insight is that employing different models will affect how well the uncertainty distribution of the calibration set covers test QA samples at a specific risk level~\citep{lin2024generating,ye2024benchmarking}, thus determining the exchangeability among the NSs. 
For instance, if the deployed model excels in health but struggles with math, the NSs from the health dataset will significantly differ from (lower than) those from the math dataset, thus leading to miscalibration, while a powerful proprietary model with comprehensive knowledge across both domains can yield an approximate uncertainty distribution. 
Furthermore, ConU methods manually remove calibration samples that fail to contain acceptable answers within the sampling space~\cite{su-etal-2024-api,kaur2024addressing,wang-etal-2024-conu,wang2024sample}, which constrains the quantity of test QA samples that the calibration set can handle, as demonstrated in Section~\ref{sec: Selective Conformal Uncertainty}. 
At this point, our goal is to derive the minimum risk level manageable by the original calibration set, and then eliminate uncertainty data outliers undermining exchangeability. 
Subsequently, the remaining test samples are expected to allow for user-specified marginal coverage. 

Inspired by prior work on outlier detection (OD) and permutation test~\citep{vovk2003testing,angelopoulos2021gentle,guan2022prediction,bates2023testing}, we propose \textbf{s}elective \textbf{con}formal \textbf{u}ncertainty (SConU), which gathers statistical evidence for nonexchangeable data sequences via hypothesis testing. 
Specifically, we construct a conformal p-value~\citep{jin2023selection,angelopoulos2024theoretical,gui2024conformal} for each test data to identify whether its uncertainty state significantly deviates from the calibration data distribution, using it as a baseline for exchangeability assessment. 
Furthermore, recognizing that uncertainty data anomalies in the calibration set compromise their reference value and statistical rigor, we provide an optimized version by incorporating the prediction status of each calibration data point at a specific risk level into the counting criterion of the conformal p-value. 
After filtering out uncertainty data outliers within the test set, we achieve rigorous management of the miscoverage rates in both single-domain and cross-domain QA datasets.

Additionally, practical NLG applications focus on conditional coverage for a particular input. 
However, this property is infeasible in most NLG cases~\citep{angelopoulos2021gentle,plassier2024probabilistic,angelopoulos2024theoretical}. 
In this paper, we investigate the impact of the exchangeability condition, split ratio, and uncertainty measurements on conditional performance, aiming to approximate conditional coverage in high-stakes QA scenarios. 
Finally, we disclose significant semantic redundancy within prediction sets in human-in-the-loop QA applications~\citep{cresswell2024conformal}.

Our contributions can be summarized as follows:
\begin{itemize}
    \item We propose \textbf{s}elective \textbf{con}formal \textbf{u}ncertainty (SConU), which for the first time implements significance tests to filter out uncertainty data outliers that violate the exchangeability precondition at a specific risk level. 
    \item We maintain the integrity of the calibration set and derive the minimum manageable risk level after deploying the language model.  
    \item We explore internal components of SConU to enhance conditional performance and operational efficiency of the prediction sets. 
\end{itemize}

%% file: Sections/RelatedWork.tex
\section{Related Work}
\label{sec: Related Work}
\paragraph{Split Conformal Prediction.} 
SCP guarantees ground-truth coverage on fresh test samples based on a calibration set~\citep{splitconformal,angelopoulos2021gentle,angelopoulos2024conformal,angelopoulos2024theoretical}. 
We briefly outline the conformal procedures of the SCP framework in Appendix~\ref{sec: SCP in classification}. 
Despite the statistical rigor, SCP assumes the NSs of all the $N$ calibration data points and the given test sample to be exchangeable~\citep{tibshirani2019conformal,bates2021distribution,barber2023conformal,farinhasnon}. Formally, the sequence of data points $Z_1, Z_2, \cdots, Z_N, Z_{N+1}$ is considered exchangeable if, for any permutation $\pi$, the sequence $\left(Z_{\pi(1)}, Z_{\pi(2)}, \dots, Z_{\pi(N)}, Z_{\pi(N+1)}\right)$ has the same joint distribution as $\left(Z_1, Z_2, \dots, Z_N, Z_{N+1}\right)$. 
Intuitively, this condition is hard to represent and verify concretely in NLG tasks~\citep{campos2024conformal}. 

\paragraph{Conformal Uncertainty in QA Tasks.} 
Recently, researchers have attempted to apply SCP to LLMs for reliable language generation. 
In white-box settings, several studies~\citep{kumar2023conformal,ye2024benchmarking,kostumov2024uncertainty,kaur2024addressing,quach2024conformal,angelopoulos2024conformal} develop ConU frameworks for multiple-choice query-answering (MCQA) and open-ended QA tasks by correlating the NS with a certain uncertainty notion of reliable responses (e.g., normalized logit-based probability of each option). 
Meanwhile, researchers also establish criteria in black-box scenarios~\citep{wang-etal-2024-conu,su-etal-2024-api,wang2024sample} based on self-consistency. 
Our work SConU applies to both settings and retains existing frameworks: We do not process calibration samples manually but instead derive the minimum risk level, which allows for handling more QA samples from diverse subjects. 
Then, we perform the conformal p-value to eliminate uncertainty data outliers violating the exchangeability precondition, and apply ConU frameworks based on the type of problems. 

Additionally, real-world QA applications often focus on conditional coverage over a particular input~\citep{gibbs2023conformal,Ding2023ClassConditionalCP,Kim2024AdaptiveUQ,cresswell2025conformal}, while in the most practical NLG case, this property is impossible to achieve~\citep{Vovk2012ConditionalVO,plassier2024probabilistic}. 
This paper examines internal factors of SConU, such as the reliability measurements in the formulation of the NS, seeking to approximate conditional coverage across various set sizes~\citep{angelopoulos2024theoretical,su-etal-2024-api,wang2024sample}. 

%% file: Sections/Method.tex
\section{Method}
\label{sec: Method}

\subsection{Preliminaries}
\label{sec: Preliminaries}
Formally, we have a held-out set of $N$ calibration data points, $\mathcal{D}_{cal}=\left\{ \left( x_i, y_i^* \right) \right\}_{i=1}^{N}$, where $x_i$ and $y_i^*$ denote the $i$-th question and ground-truth answer, respectively. 
For each data point, we sample multiple (e.g., $M$) responses from the output space of the language model to construct a candidate set for the corresponding question, denoted as $\left\{ y_j^{(i)} \right\}_{j=1}^{M}$.\\
We can calculate the reliability score of each generation or semantic cluster utilizing various uncertainty measurements within the candidate set~\cite{su-etal-2024-api,wang-etal-2024-conu,kaur2024addressing}. 
For instance, we can express the confidence score of each option in MCQA task as $w_l \cdot F_l \left(y_j^{(i)}\right)+w_f \cdot F_f \left(y_j^{(i)}\right)$, where $F_l \left(y_j^{(i)}\right)$ represents the probability derived from model logit, $F_f \left(y_j^{(i)}\right)$ denotes the frequency score of $y_j^{(i)}$ within the candidate set, and $w_l$ and $w_f$ are the respective weights assigned to each score. 
Then, the NS of each MCQA sample is $1 - w_l \cdot F_l \left(y_i^*\right) - w_f \cdot F_f \left(y_i^*\right)$ ($w_l+w_f=1$). 

Due to the randomness of sampling and potential limitations in model capability, we may not always obtain an acceptable response that aligns with the ground-truth answer by sampling $M$ times for each QA sample. 
Unlike prior work~\cite{wang-etal-2024-conu,kaur2024addressing,wang2024sample}, we do not demand that samples employed as the calibration data must encompass acceptable responses within their candidate sets. 
On one hand, given that SCP is model-agnostic, we cannot guarantee that all employed language models in practical applications will be capable of addressing the same questions. 
Furthermore, we aim for the calibration set to cover data distributions across various domains comprehensively. 
While the lower bound of the error rate that the calibration set can control is constrained at this point, we can accommodate a greater volume of test QA samples by easing the risk level of $\alpha$. 

In the following section, (1) we first introduce our two developed conformal p-values that assess the exchangeability condition through significance tests. Then, (2) we formally verify the necessity of maintaining the integrity of the original calibration set. Next, (3) we investigate the minimum risk level manageable by the original calibration set. Finally, (4) we present the workflow of our framework.

\subsection{Selective Conformal Uncertainty}
\label{sec: Selective Conformal Uncertainty}
Inspired by prior research~\cite{permutation1937,jin2023selection,bates2023testing,gui2024conformal,angelopoulos2024theoretical}, we collect statistical evidence for nonexchangeable sequences of NSs arising from uncertainty data outliers via hypothesis testing. 
Specifically, we define the null hypothesis $\mathcal{H}_0$ for the test data point $x_{N+1}$ with the significance level of $\delta$ as follows: $\left\{ \left( x_i, y_i^* \right) \right\}_{i=1}^{N}$ can serve as the calibration set for $x_{N+1}$ with coverage guarantees. 
Rejecting $\mathcal{H}_0$ indicates sufficient evidence of the prediction set with an unbounded miscoverage rate when tackling $x_{N+1}$ based on $\left\{ \left( x_i, y_i^* \right) \right\}_{i=1}^{N}$. 
To this end, we construct a finite-sample valid conformal p-value associating $\mathcal{H}_0$ as
\begin{equation}\label{eq: baseline_sconu}
        p_{N+1} = \frac{1+\textstyle\sum_{i=1}^{N}\textbf{1} \left \{ u_i \geq u_{N+1}\right \}}{N+1}
\end{equation}

In the formulation, $u_i$ indicates the uncertainty of the language model addressing the $i$-th question $x_i$, measured by an uncertainty notion $U$, and we utilize the predictive entropy (PE)~\citep{kadavath2022language,duan-etal-2024-shifting,WANG2025109553}. 
Note that the uncertainty corresponds to the output distribution of a particular QA sample, while the NS reflects the model's uncertainty regarding a specific generation, representing the disagreement between the current response and the query.

As mentioned, we consider that uncertainty data anomalies may present in the calibration set and compromise statistical rigor. 
To examine the reference quality of each calibration data point at a specific risk level, we refine the conformal p-value: 
\begin{equation}\label{eq: enhanced_sconu}
    \begin{split}
        &p^{'}_{N+1} = \\ &\frac{1+\textstyle\sum_{i=1}^{N}\textbf{1} \left \{ u_i \geq u_{N+1} , y_i^* \in E\left( x_i, \mathcal{D}_{cal}, \alpha \right) \right \}}{N+1},
    \end{split}
\end{equation}
where $y_i^* \in E\left( x_i, \mathcal{D}_{cal}, \alpha \right)$ determines whether the prediction set established for $x_i$, calibrated by all data points in $\mathcal{D}_{cal}$ except for $x_i$, contains $y_i^*$ at a risk level of $\alpha$. 
If not, we intuitively consider that the model may encounter hallucination issues when processing $x_i$~\citep{kuhn2023semantic,farquhar2024detecting}, or that the uncertainty of its output distribution is abnormally high, which results in high NSs of its reliable generations and miscoverage. 
At this point, $u_i \geq u_{N+1}$ lacks statistical validity at the risk level of $\alpha$, and $\mathbf{1}\left\{\cdot\right\}$ does not count. 

For simplicity, we refer to the conformal procedure employing two conformal p-values as SConU and SConU-Pro in the following text. 
We demonstrate that the two conformal p-values adhere to the statistical definition of p-values in Appendix~\ref{sec: appendix Conformal p-value}, and present a more rigorous framework to detect when test points do not come from the same distribution. 

\paragraph{Maintenance of the calibration set.} 
As mentioned, we do not remove calibration data that fail to cover acceptable responses within their candidate sets. 
In this section, we demonstrate the practical significance by defining the minimum sampling size of each calibration QA sample as 
\begin{equation}
    m_i = \inf \left\{ M_i: \forall M_i^{'} \geq M_i, y_i^* \in \left\{ y_j^{(i)} \right\}_{j=1}^{M_i^{'}}\right\},
\end{equation}
which ensures that there is at least one correct answer in the $i$-th candidate set of size $m_i$. 
Then, we sort the $N$ minimum sampling sizes and calculate their $\frac{\left \lceil \left ( 1-\beta\right ) \left ( 1+N \right ) \right \rceil}{N}$ quantile: $\hat m = m_{\left \lceil \left ( 1-\beta\right ) \left ( 1+N \right ) \right \rceil}$, where $\beta$ represents the error rate (similar to $\alpha$). 
If the test sample is exchangeable with $N$ calibration data points, we have $\mathbb{P}\left(m_{N+1} \leq m_i\right)=\frac{i}{N+1}$. 
We then set the sampling size of the test QA sample to $\hat{m}$ and obtain the probability of covering at least one admissible response within $\hat{m}$ sampling times
\begin{equation}\label{eq: sampling size calibration probability}
    \begin{split}
        &\mathbb{P}\left ( y^*_{N+1} \in \left\{ y_j^{(N+1)} \right\}_{j=1}^{\hat{m}} \right ) = \mathbb{P}\left( m_{N+1} \leq \hat{m}\right)\\
        & = \frac{\left \lceil \left ( 1-\beta\right ) \left ( 1+N \right ) \right \rceil}{N+1} \geq 1-\beta
    \end{split},
\end{equation}

Following the requirement of previous research, where at least one correct answer exists in the candidate set of fixed size $M$ for each calibration data, we have $M \geq \max \left\{m_i\right\}_{i=1}^{N}$ and $\beta \rightarrow 0$.  
At this point, $y^*_{N+1} \in \left\{ y_j^{(N+1)} \right\}_{j=1}^{M}$ is a certain event, which is infeasible in practical NLG tasks. 
Additionally, removing calibration samples will narrow the uncertainty distribution of the calibration set, which diminishes its adaptability to new test QA samples. 
Therefore, we explore the minimum risk level controlled by the original calibration set. 

\paragraph{Minimum risk level.} 
Building on prior research \citep{angelopoulos2024conformal,farinhasnon}, we post-process the candidate set of each calibration data point into a set of reliable responses with sufficiently high confidence scores, $\mathcal{C}_{\lambda}\left(x_i\right)=\left\{ y_j^{(i)}: F\left(y_j^{(i)}\right) \geq 1-\lambda \right\}$ ($\lambda \in \left[0, 1\right]$), where $F\left(\cdot\right)$ can be any measurement that reflect the trustworthiness of each sampled response. 
Then, we calculate the loss of miscoverage, $l\left( \mathcal{C}_{\lambda}\left( x_i \right), y_i^* \right) = \mathbf{1}\left\{ y_i^* \notin \mathcal{C}_{\lambda}\left( x_i \right) \right\}$, abbreviated as $l_i\left(\lambda\right)$, and set $L_N\left( \lambda \right)=\frac{1}{N} \displaystyle\sum_{i=1}^{N} l_i\left(\lambda\right)$. 
Suppose $l_{N+1}\left(\lambda\right)$ follows $\mathrm{Uniform}\left( \left\{ l_1\left(\lambda\right),\cdots,l_{N+1}\left(\lambda\right) \right\} \right)$ by exchangeability, we have $\mathbb{E}\left[ l_{N+1}\left(\lambda\right) \right] = \frac{1}{N+1} \displaystyle\sum_{i=1}^{N+1} l_i\left(\lambda\right) = \frac{N L_N\left( \lambda \right)+l_{N+1}\left(\lambda\right)}{N+1}$. 
Obviously, $L_N\left( \lambda \right)$ is non-increasing in $\lambda$. 
Then, we set $\lambda$ to its upper bound (i.e., 1) and obtain the minimum value, $L_N\left( 1 \right)$. 

When $\lambda$ is set to 1, $\mathcal{C}_{\lambda}\left( x_i \right)=\left\{ y_j^{(i)} \right\}_{j=1}^{M}$, and at this point, the problem simplifies to calculating the proportion of candidate sets in the calibration set that do not contain an acceptable response: $L_N\left( 1 \right)=\frac{1}{N} \displaystyle\sum_{i=1}^{N} \mathbf{1}\left\{ y_i^* \notin \left\{ y_j^{(i)} \right\}_{j=1}^{M} \right\}$. 
Since $\mathbb{E}\left[ l_{N+1}\left(\lambda\right) \right]$ should be controlled by a user-specified risk level of $\alpha$ (i.e., $\mathbb{E}\left[ l_{N+1}\left(\lambda\right) \right]\leq \alpha$), and $l_{N+1}\left(\lambda\right) \in \left\{0, 1\right\}$, we obtain $\mathbb{E}\left[ l_{N+1}\left(\lambda\right) \right] \geq \frac{N L_N\left( 1 \right)}{N+1}$, and at this point, 
\begin{equation}\label{eq: minimum risk level}
    \alpha_l= N L_N\left( 1 \right) / \left(N+1\right)
\end{equation}
Finally, for any risk level of $\alpha \geq \alpha_l$, we can rigorously manage the correctness miscoverage rate leveraging the given calibration set. 

The concept of minimum risk level also aligns with abstention~\cite{yadkori2024mitigating,shahrokhi2025conformal}. 
Calibration methods operating with finite-sampling are constrained by the LLM’s capacity to generate admissible outputs within this finite horizon. 
For $\alpha < \alpha_l$, we have to enumerate the entire output space to maintain valid coverage on some inputs, and in such cases, the prediction set will not provide practical information. 

\begin{figure}
    \centering
    \includegraphics[width=\linewidth]{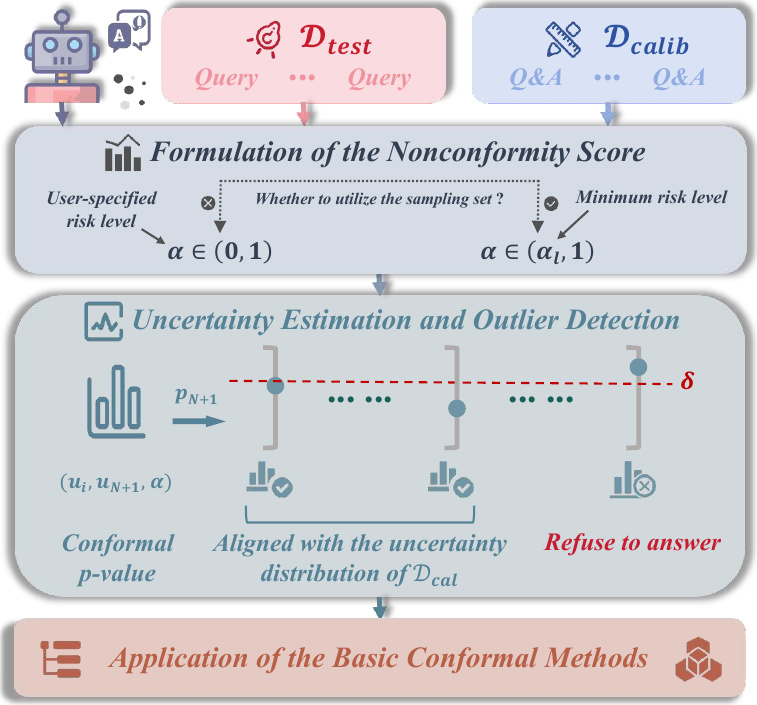}
    \caption{Pipeline of the SConU framework. 
    We achieve rigorous coverage of correct generations on test samples at various user-specified risk levels based on the calibration set after automatic outliers detection.}
    \label{fig: pipeline}
    \vspace{-5mm}
\end{figure}

\paragraph{Workflow of SConU.} 
As illustrated in Figure~\ref{fig: pipeline}, after deploying the LLM and the maintained calibration set, we first calculate the minimum risk level $\alpha_l$ if we utilize the sampling set when formulating NS; otherwise, we allow any user-specified risk level $\alpha \in (0,1)$. 
Then, given each test sample, we conduct significance tests to identify whether it aligns with the uncertainty distribution of the calibration set at the risk level of $\alpha$. 
A low conformal p-value suggests a violation of the exchangeability precondition, and we decline to respond. 
After filtering out outliers, we conduct conformal procedures for samples within the remaining test set with finite-sample guarantees of correctness coverage. 


%% file: Sections/Experiments.tex
\input{Tables/sampling_size_calibration}
\input{Tables/Single_SConU}
\section{Experiments}
\label{sec: Experiments}
\subsection{Experimental Settings}
\paragraph{Datasets.}
We utilize 3 closed-ended QA datasets: MMLU~\citep{hendrycks2021measuring} for multitask language understanding, more challenging MMLU-Pro~\citep{wang2024mmlu}, and MedMCQA~\citep{pal2022medmcqa} for real-world medical entrance exam, and 2 open-domain datasets: TriviaQA~\citep{joshi2017triviaqa} for closed-book QA and CoQA~\citep{reddy2019coqa} for open-book conversational QA. 
More details are presented in Appendix~\ref{sec: Details of Datasets}.

\paragraph{Metrics.} 
We utilize the Empirical Miscoverage Rate (EMR) to assess whether conformal methods produce prediction sets that meet statistical guarantees~\citep{wang2024sample,quach2024conformal} after outlier elimination. 
For conditional coverage, we apply the Size-stratified Miscoverage Rate (SMR) that evaluates error rates across various set sizes~\citep{angelopoulos2021gentle,kumar2023conformal,su-etal-2024-api}. 
We also explore the operational efficiency through the Average Prediction Set Size (APSS) on the test set~\citep{wang-etal-2024-conu,wang2024sample,su-etal-2024-api,angelopoulos2024theoretical}. 

Our utilized LLMs and additional experimental settings are presented in Appendix~\ref{sec: Additional Experimental Settings}. 

\begin{figure*}[!t]
    \centering
     \begin{subfigure}{0.495\linewidth}
        \centering
        \includegraphics[width=\linewidth]{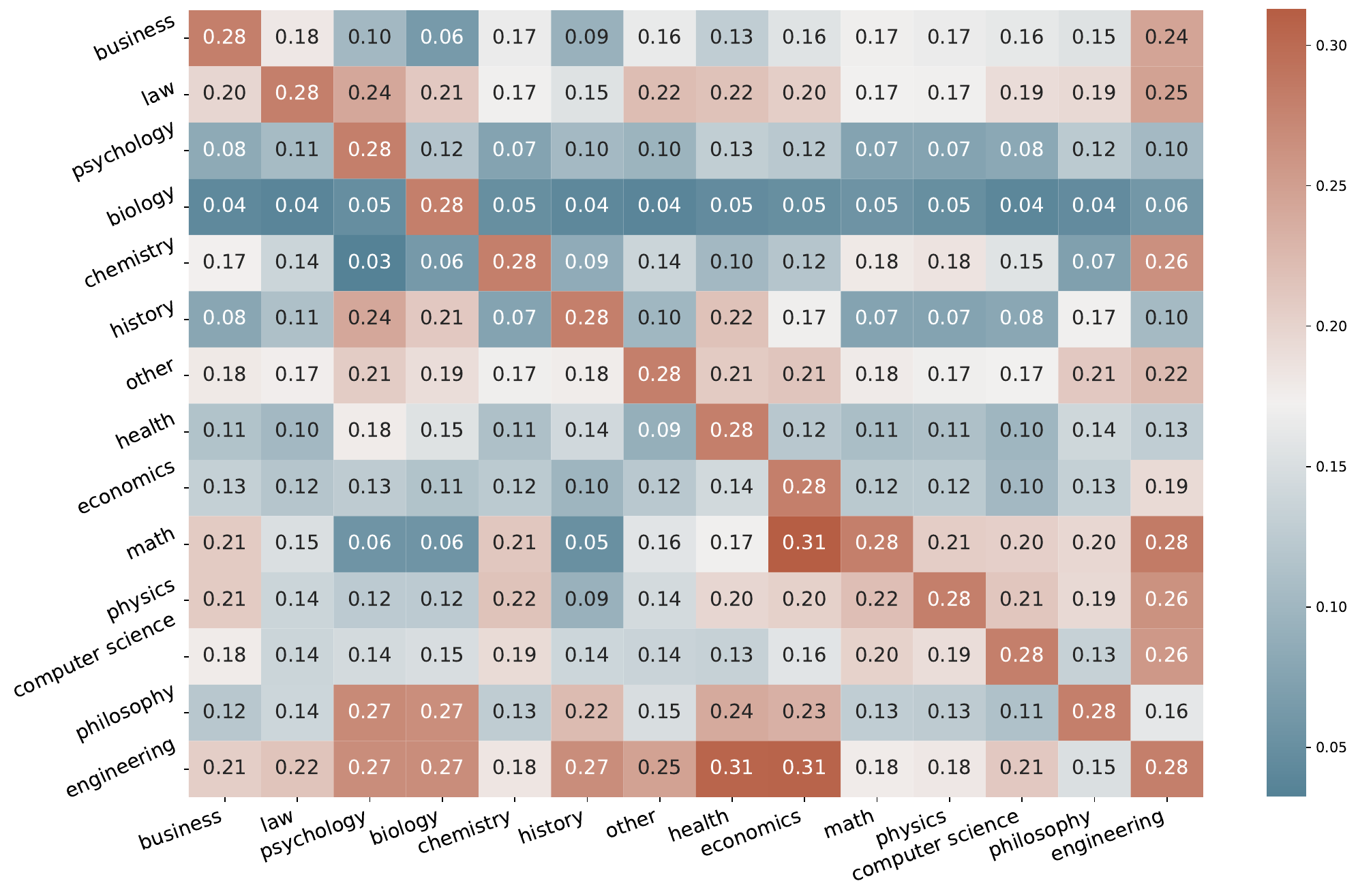}
        \caption{EMR results of SConU.}
	\label{fig: cross-domain baseline}
    \end{subfigure}
    \hfill
    \centering
    \begin{subfigure}{0.495\linewidth}
	\centering
	\includegraphics[width=\linewidth]{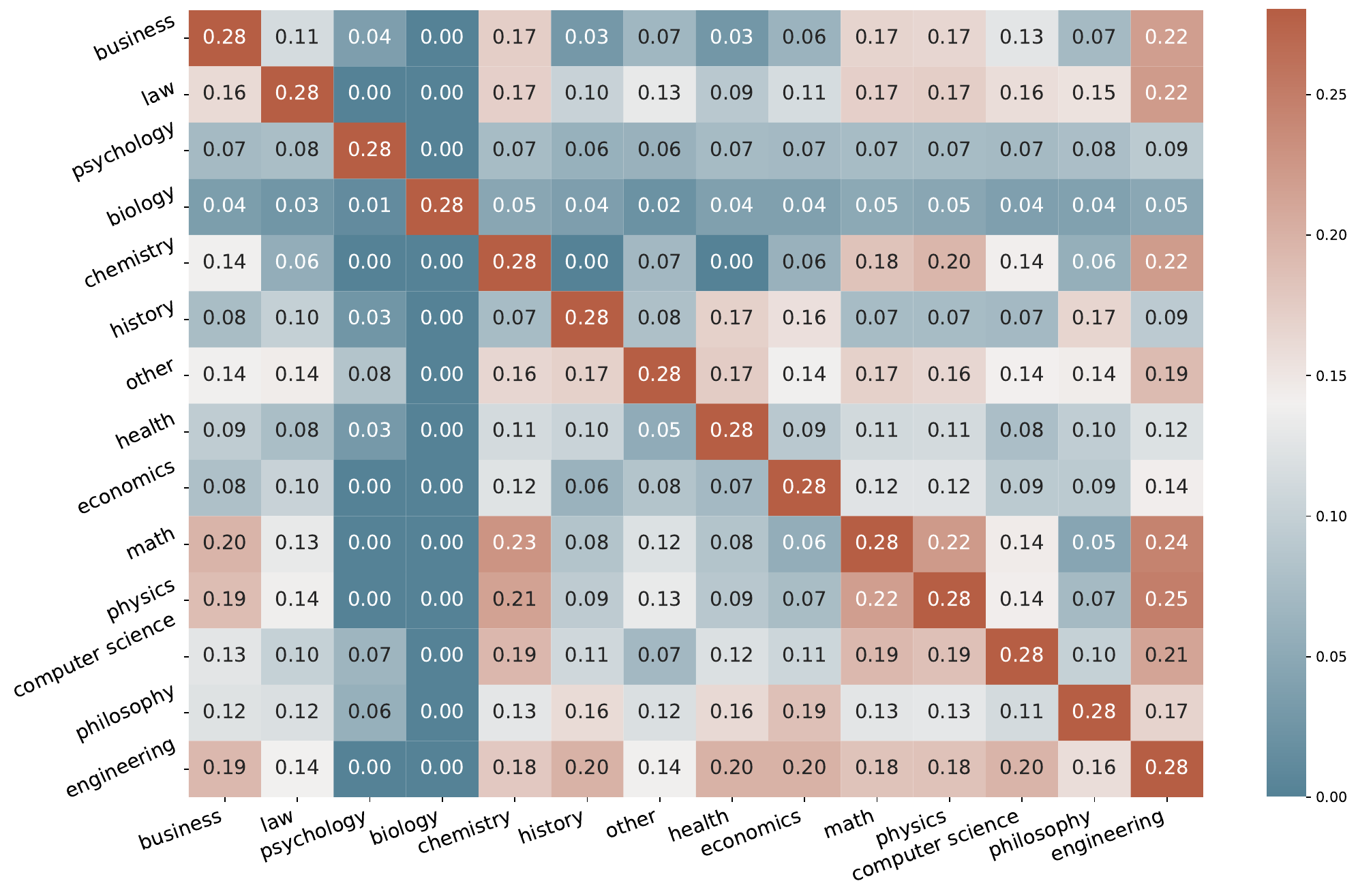}
	\caption{EMR results of SConU-Pro.}
	\label{fig: cross-domain sconu}
    \end{subfigure}
    \caption{Results of EMR after applying our two frameworks utilizing the LLaMA-3.1-8B-Instruct model on the MMLU-Pro dataset. Note that all data on the diagonal is manually set to equal to $\alpha$ ($\alpha_l=0.2723$).}\label{fig: cross-domain comparsion}
\end{figure*}

\begin{figure*}[!t]
    \centering
     \begin{subfigure}{0.495\linewidth}
        \centering
        \includegraphics[width=\linewidth]{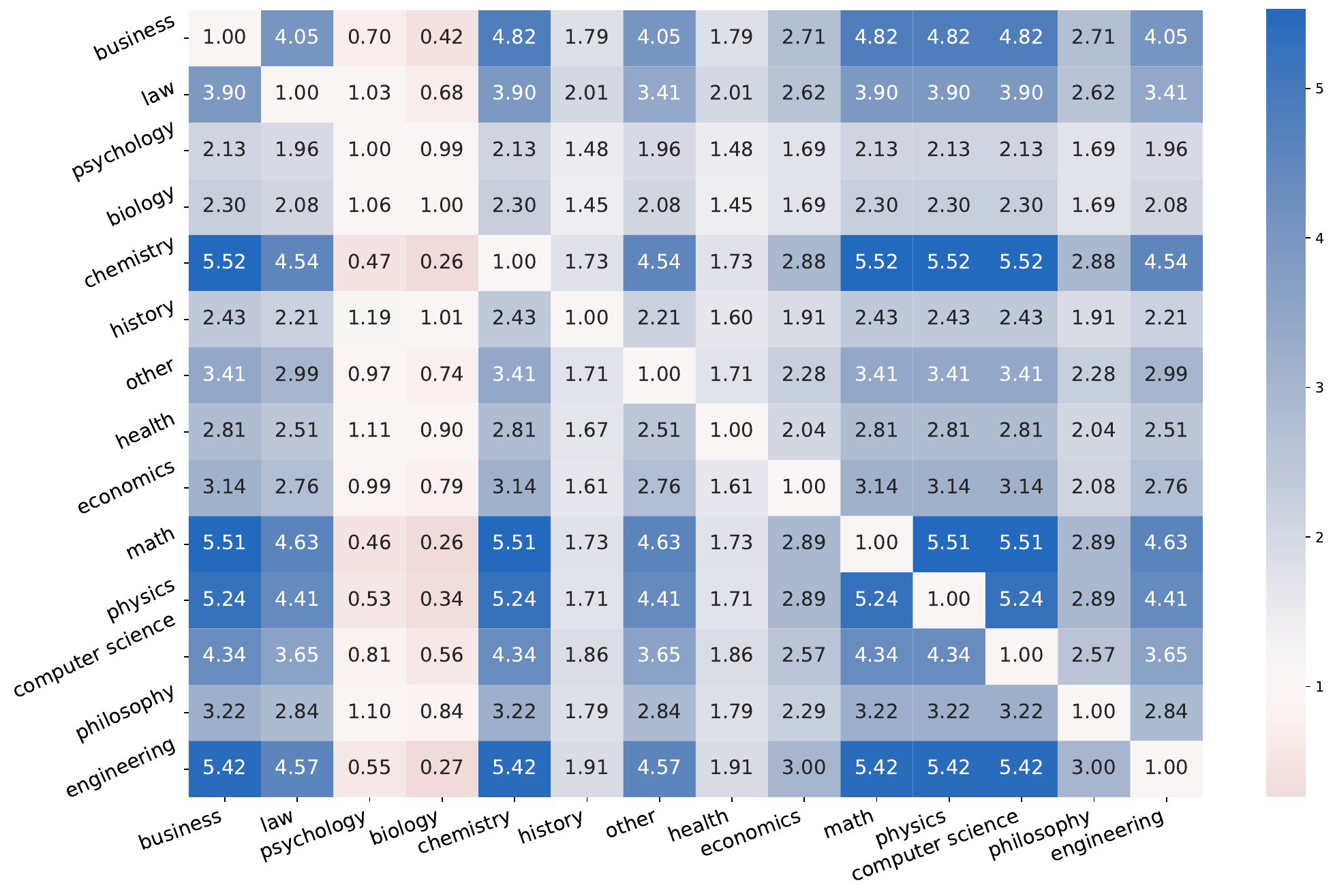}
        \caption{Original APSS results.}
	\label{fig: original APSS}
    \end{subfigure}
    \hfill
    \centering
    \begin{subfigure}{0.495\linewidth}
	\centering
	\includegraphics[width=\linewidth]{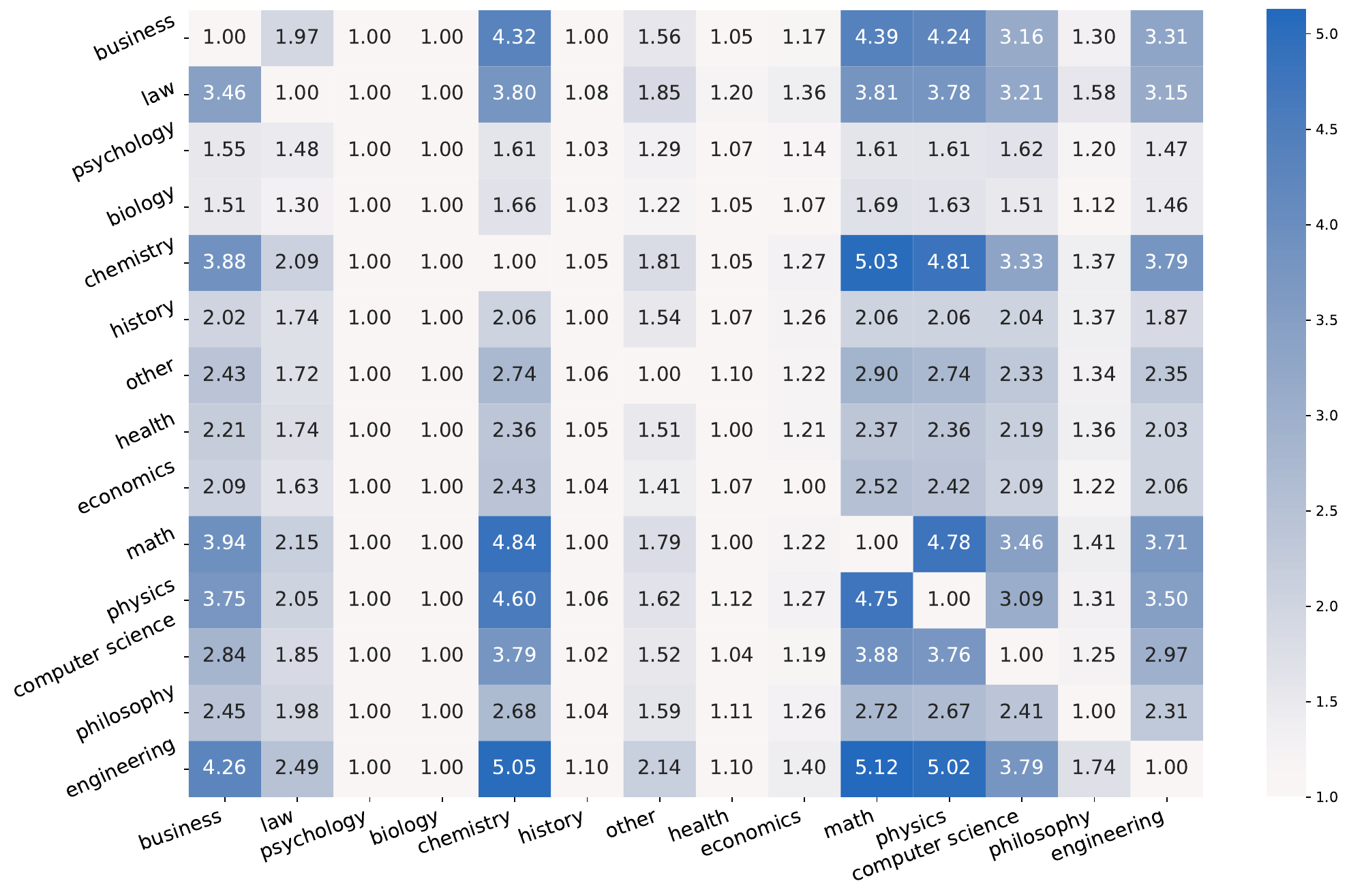}
	\caption{APSS results of SConU-Pro.}
	\label{fig: sconu-pro APSS}
    \end{subfigure}
    \caption{Results of APSS before and after performing SConU-Pro, utilizing the LLaMA-3.1-8B-Instruct model on the MMLU-Pro dataset. Note that all data on the diagonal is manually set to 1.}\label{fig: APSS comparsion}
\end{figure*}

\subsection{Empirical Results}
\label{sec: Empirical Results}
\paragraph{Calibration of Sampling Size.} 
As theoretically demonstrated in Section~\ref{sec: Selective Conformal Uncertainty}, we maintain the integrity of the calibration set to accommodate more test samples. 
Here, we empirically validate the practicality of Eq.~\eqref{eq: sampling size calibration probability}. 
We apply the sampling size calibration procedures to both open-ended TriviaQA and closed-ended MedMCQA tasks using four LLMs. 
For each setting, we randomly split the calibration and test sets 100 times with a 0.5 split ratio. 
We determine the minimum sampling size for each test data point based on the calibration set and a user-specified risk level, $\beta$. 
As presented in Table~\ref{table: sampling size calibration}, the average miscoverage rate is rigorously bounded (i.e., $\leq$) by $\beta$, which underscores the importance of maintaining the integrity of the calibration set under exchangeable conditions.

\paragraph{Marginal Coverage.} 
As illustrated in Figure~\ref{fig: nonexchangeable-single}, we apply ConU to single-domain datasets and observe that the mean EMR results exceed the user-specified risk levels for some LLMs (e.g., Qwen-2-7B-Instruct). 
Moreover, the shaded area significantly surpassing the dashed line indicates substantial issues unbounded EMR in 100 trials. 
Note that we employ the typical conformal framework~\citep{kumar2023conformal,ye2024benchmarking,kostumov2024uncertainty,campos2024conformal} for MCQA tasks, detailed in Appendix~\ref{sec: Details of Utilizing the Conformal Frameworks}. 
We implement our SConU framework under the same settings using the Qwen-2-7B-Instruct and LLaMa-3.1-8B-Instruct models as examples. 
We also consider the median metric as mentioned in several studies~\citep{deng2023distributionfree,snell2023quantile,zollo2024prompt}. 
As shown in Table~\ref{table: single-domain sconu}, both the mean and median of the EMR results obtained from SConU are rigorously confined within the risk level, and the variance metric is significantly lower than that of the basic ConU framework on the Health and Economics subsets, highlighting the effectiveness of our approach. 

\input{Tables/Conditional_ablation}
\input{Tables/Vicuna_conditional}

In real-world QA tasks, LLMs often face queries from diverse disciplines~\citep{kumar2023conformal}. 
However, as shown in Figure~\ref{fig: nonexchangeable-cross}, considerable issues of unbounded EMR emerge when the uncertainty distribution of test samples deviates from that of the provided calibration set, compromising the reliability of their prediction sets. 
For instance, when utilizing calibration data from the Psychology domain to address test samples from 13 other subsets, EMR values typically exceed the risk level of 0.28, peaking at 0.83 in the Math subject. 
Moreover, we may have no access to model logit. 
At this point, we incorporate the frequency score into the NS formulation and set $w_l=0, w_f=1$ following the study~\citep{wang2024sample}. 
Then, we employ our SConU framework, which filters out uncertainty data outliers within each test subset. 
As illustrated in Figure~\ref{fig: cross-domain baseline}, the EMR metric for the Math discipline decreases to 0.15, while the results for other subjects remain confined by the minimum risk level of 0.28. 
When subsets from other disciplines are utilized as the calibration set, EMR results generally meet the guarantee of marginal coverage.

Despite the theoretical guarantee of SCP being rigorous, there can be minor fluctuations in practice due to finite-sample variability~\citep{angelopoulos2021gentle,ye2024benchmarking,angelopoulos2024theoretical}. 
We notice EMR deviations in the results of SConU. 
To address this, we apply SConU-Pro by incorporating the prediction status of each calibration data point into the counting criteria of the conformal p-value, which evaluates the reference values of the calibration samples across various risk levels. 
As demonstrated in Figure~\ref{fig: cross-domain sconu}, we achieve rigorous management of the EMR metric (i.e., $\leq \alpha$) in cross-domain scenarios. 
Furthermore, we compare the APSS metric before and after implementing SConU-Pro. 
As illustrated in Figure~\ref{fig: APSS comparsion}, when employing the Psychology or Biology subset as the calibration set, we observe APSS being less than 1 in the test sets of other disciplines, indicating that many test QA samples have empty prediction sets. 
Following the application of SConU-Pro, we attain an APSS metric of 1 for all selected test samples with the majority of EMR metrics equal to 0, suggesting that we accurately identify the correct answer for each test QA sample. 
In other calibration settings, the APSS results also exhibit a significant decline, thereby enhancing prediction efficiency. 

More details of our performed conformal procedures can be found in Appendix~\ref{sec: Details of Utilizing the Conformal Frameworks}, and additional experimental results are presented in Appendix~\ref{sec: additional result in appendix}. 

\input{Tables/Semantic_redundancy}

\paragraph{Conditional Coverage.} 
Given the critical importance of correctness coverage for individual samples in high-stakes QA tasks, we explore four key factors: exchangeability, the reliability of the NS in representing disagreements between query-answer pairs, split ratio, and model performance, and examine EMR across various set sizes. 
Our analysis focuses on the MedMCQA task and the Clinical Knowledge subset of the MMLU dataset. 
As presented in Table~\ref{table: conditional ablation study}, when employing logit-based NSs, EMR values exceed the risk threshold at set sizes of 1 and 3. 
By incorporating the frequency score into the NS formulation and appropriately increasing the sample size, we observe a reduction in the SSM metric. 
Moreover, while more calibration samples enhance conditional performance, the SSM metric remains above the acceptable risk level. 
To address this, we utilize the conformal p-value to eliminate outliers, achieving approximate conditional coverage, with the SSM metric falling below the risk threshold at both split ratios. 
For instance, with a split ratio of 0.5, we attain an SSM value of 0.3092 using the frequency score derived from the candidate set of size 10. 
As shown in Table~\ref{table: vicuna7_13 conditional coverage}, model performance also plays a significant role in influencing conditional coverage, and our SConU-Pro framework consistently enhances the SSM metric. 

We conclude that we can design NS using more reliable uncertainty measures based on the internal model information and the true sampling distribution. 
Additionally, we can appropriately increase the scale of the calibration data, although this will increase computational costs. 
Most importantly, it is essential to ensure exchangeability among QA samples. 
Finally, deploying task-specific models can further improve conditional performance.

\paragraph{Prediction Efficiency.}
In open-domain QA tasks, we observe significant semantic redundancy in the prediction sets generated by previous ConU frameworks~\citep{wang-etal-2024-conu,su-etal-2024-api}. 
As shown in Table~\ref{table: semantic redundancy}, the mean APSS from 100 trials decreases markedly before and after semantic deduplication, suggesting that there is considerable potential for improving the action efficiency of these prediction sets while maintaining the guarantee.


%% file: Tables/sampling_size_calibration.tex
\begin{table*}[t!]
\centering
\caption{Results of the probability (mean ± std) of failing to obtain admissible responses through calibrating the sampling size for each question within the test set across various values of $\beta$.}\label{table: sampling size calibration}
\adjustbox{max width=\linewidth}{
    \begin{tabular}{ccccccc} 
        \toprule
        
        \textbf{Dataset} & \multicolumn{3}{c}{\textbf{TriviaQA (open-ended)}} &  \multicolumn{3}{c}{\textbf{MedMCQA (closed-ended)}}\\
        
        \midrule

        \textbf{LLMs} $/ \beta$ & \textbf{0.1} & \textbf{0.2} & \textbf{0.3} & \textbf{0.1} & \textbf{0.2} & \textbf{0.3}\\

        LLaMA-3.2-3B-Instruct & 0.0884 ± 0.0149 & 0.1767 ± 0.0109 & 0.2725 ± 0.0194 & 0.0896 ± 0.0078 & 0.1823 ± 0.0084 & 0.2423 ± 0.0072\\

        OpenChat-3.5 & 0.0848 ± 0.0179 & 0.1551 ± 0.0391 & 0.1997 ± 0.0090 & 0.0911 ± 0.0119 & 0.1785 ± 0.0265 & 0.2676 ± 0.0074\\

        LLaMA-3.1B-Instruct & 0.0869 ± 0.0060 & 0.1770 ± 0.0378 & 0.1965 ± 0.0086 & 0.0861 ± 0.0067 & 0.1697 ± 0.0331 & 0.2771 ± 0.0078\\

        Qwen2.5-14B-Instruct & 0.0835 ± 0.0201 & 0.1731 ± 0.0075 & 0.1731 ± 0.0075 & 0.0815 ± 0.0047 & 0.0815 ± 0.0047 & 0.0815 ± 0.0047\\
        
        \bottomrule
    \end{tabular}
}
\end{table*}

%% file: Tables/Single_SConU.tex
\begin{table*}[t!]
\centering
\caption{The EMR results obtained from 100 trials on the MMLU-Pro dataset. \textbf{Note that} the mean and median metrics only need to be below the corresponding risk level, and they are not required to be as low as possible. \faTimesCircle $\,$ indicates using the basic ConU framework, and \faCheckCircle $\,$ represents utilizing our SConU criterion, eliminating uncertainty data outliers within the test set. {\color{customred} Red} indicates violation of the risk level.}\label{table: single-domain sconu}
\adjustbox{max width=\linewidth}{
    \begin{tabular}{cccccccccccc} 
        \toprule
        
        \textbf{Disciplinary} & \textbf{Metric} & \textbf{OD} & \textbf{0.1} & \textbf{0.2} & \textbf{0.3} & \textbf{0.4} & \textbf{0.5} & \textbf{0.6} & \textbf{0.7} & \textbf{0.8} & \textbf{0.9} \\
        
        \midrule
        
        \multicolumn{12}{c}{Qwen-2-7B-Instruct Model.}\\
        
        \hline

        \multirow{6}{*}{Health} & \multirow{2}{*}{Mean} & \faTimesCircle & {\color{customred} 0.1019} & 0.1977 & {\color{customred} 0.3001} & {\color{customred} 0.4035} & {\color{customred} 0.5004} & 0.5964 & 0.6888 & 0.7938 & 0.8788\\
        {} & {} &  \faCheckCircle &  0.0938 &  0.1943 &  0.2972 &  0.3957 &  0.4937 &  0.5915 &  0.6819 &  0.7876 &  0.8754\\
        {} & \multirow{2}{*}{Std $\downarrow$} & \faTimesCircle & 0.0285 & 0.0372 & 0.0420 & 0.0434 & 0.0424 & 0.0441 & 0.0420 & 0.0323 & 0.0232\\
        {} & {} &  \faCheckCircle &  \textbf{0.0283} &  \textbf{0.0362} &  0.0423 &  \textbf{0.0425} &  \textbf{0.0358} &  \textbf{0.0429} &  \textbf{0.0384} &  \textbf{0.0323} &  \textbf{0.0227}\\
        {} & \multirow{2}{*}{Median} & \faTimesCircle & {\color{customred} 0.1080} & 0.1960 & 0.2960 & {\color{customred} 0.4120} & {\color{customred} 0.5080} & 0.5960 & 0.6760 & 0.7880 & 0.8760\\
        {} & {} &  \faCheckCircle &  0.0960 &  0.1920 &  0.2920 &  0.3960 &  0.4920 &  0.5920 &  0.6800 &  0.7960 &  0.8800\\

        
        \hline

        \multirow{6}{*}{Economics} & \multirow{2}{*}{Mean} & \faTimesCircle & {\color{customred} 0.1001} & {\color{customred} 0.2032} & 0.2951 & 0.3928 & 0.4916 & 0.5871 & 0.6838 & 0.7658 & 0.8783\\
        {} & {} &  \faCheckCircle &  0.0965 &  0.1951 &  0.2950 &  0.3928 &  0.4877 &  0.5853 &  0.6820 &  0.7630 &  0.8767\\
        {} & \multirow{2}{*}{Std $\downarrow$} & \faTimesCircle & 0.0279 & 0.0338 & 0.0367 & 0.0408 & 0.0384 & 0.0366 & 0.0347 & 0.0352 & 0.0161\\
        {} & {} &  \faCheckCircle &  \textbf{0.0210} &  \textbf{0.0281} &  \textbf{0.0275} &  \textbf{0.0395} &  \textbf{0.0294} &  \textbf{0.0253} &  \textbf{0.0294} &  \textbf{0.0272} &  0.0226 \\
        {} & \multirow{2}{*}{Median} & \faTimesCircle & {\color{customred} 0.1040} & {\color{customred} 0.2080} & 0.2880 & 0.3920 & 0.4960 & 0.5920 & 0.6880 & 0.7640 & 0.8760\\
        {} & {} &  \faCheckCircle &  0.0960 &  0.1960 &  0.2920 &  0.3960 &  0.4880 &  0.5880 &  0.6840 &  0.7680 &  0.8720\\


        \hline
        
        \multicolumn{12}{c}{LLaMA-3.1-8B-Instruct Model.}\\
        
        \hline

        \multirow{6}{*}{Health} & \multirow{2}{*}{Mean} & \faTimesCircle & 0.0961 & 0.1933 & 0.2922 & 0.3912 & 0.4957 & 0.5988 & 0.6936 & {\color{customred} 0.8028} & {\color{customred} 0.9015}\\
        {} & {} &  \faCheckCircle &  0.0975 &  0.1925 &  0.2926 &  0.3935 &  0.4978 &  0.5966 &  0.6883 &  0.7913 &  0.8941\\
        {} & \multirow{2}{*}{Std $\downarrow$} & \faTimesCircle & 0.0273 & 0.0364 & 0.0459 & 0.0447 & 0.0481 & 0.0459 & 0.0404 & 0.0362 & 0.0257\\
        {} & {} &  \faCheckCircle &  \textbf{0.0214} &  \textbf{0.0300} &  \textbf{0.0412} &  \textbf{0.0431} &  \textbf{0.0457} &  \textbf{0.0420} &  0.0426 &  \textbf{0.0357} &  \textbf{0.0241}\\
        {} & \multirow{2}{*}{Median} & \faTimesCircle & 0.0960 & 0.1920 & 0.2960 & 0.3920 & 0.4960 & 0.5880 & 0.6920 & 0.7960 & {\color{customred} 0.9040}\\
        {} & {} &  \faCheckCircle &  0.0960 &  0.1960 &  0.3000 &  0.3880 &  0.4840 &  0.5920 &  0.6960 &  0.7960 &  0.8920\\
        
        
        \hline

        \multirow{6}{*}{Economics} & \multirow{2}{*}{Mean} & \faTimesCircle & 0.0947 & 0.1952 & 0.2997 & {\color{customred} 0.4018} & 0.4985 & 0.5932 & 0.6936 & 0.7889 & 0.8867\\
        {} & {} &  \faCheckCircle &  0.0916 &  0.1902 &  0.2913 &  0.3875 &  0.4855 &  0.5879 &  0.6855 &  0.7897 &  0.8863\\
        {} & \multirow{2}{*}{Std $\downarrow$} & \faTimesCircle & 0.0363 & 0.0373 & 0.0424 & 0.0443 & 0.0458 & 0.0447 & 0.0385 & 0.0326 & 0.0279\\
        {} & {} &  \faCheckCircle &  \textbf{0.0242} &  \textbf{0.0368} &  \textbf{0.0415} &  \textbf{0.0427} &  \textbf{0.0455} &  \textbf{0.0388} &  \textbf{0.0294} &  \textbf{0.0285} &  \textbf{0.0250}\\
        {} & \multirow{2}{*}{Median} & \faTimesCircle & 0.1000 & 0.1880 & 0.2920 & {\color{customred} 0.4080} & 0.4880 & 0.5840 & 0.6800 & 0.7880 & 0.8840\\
        {} & {} &  \faCheckCircle &  0.0920 &  0.1920 &  0.2920 &  0.3960 &  0.4880 &  0.5960 &  0.6920 &  0.7920 &  0.8880\\


        \bottomrule
    \end{tabular}
}
\end{table*}

%% file: Tables/Conditional_ablation.tex
\begin{table*}[t!]
\centering
\caption{Results of the SSM metric obtained from 100 trials, under different settings on the MedMCQA dataset, utilizing the Qwen-2.5-14B-Instruct model (Mean $\pm$ Std). {\color{customred} Red} indicates violation of the risk level.}\label{table: conditional ablation study}
\adjustbox{max width=\linewidth}{
    \begin{tabular}{cccccccc} 
        \toprule
        
        \textbf{$w_l$ (Logit)} & \textbf{$w_f$ (Frequency)} & \textbf{$M$ (Sampling)} & \textbf{OD} & \textbf{Size $=$ 1} & \textbf{Size $=$ 2} & \textbf{Size $=$ 3} & \textbf{SSM} $\downarrow$\\
        \midrule
        \multicolumn{8}{c}{Split ratio is fix at 0.5 and $\alpha$ is set to 0.34 ($\alpha_l=0.3342$).}\\
        \hline
        1 & 0 & 10 & \faTimesCircle & {\color{customred} 0.3428 $\pm$ 0.0151} & 0.2800 $\pm$ 0.0277 & 0.1056 $\pm$ 0.1860 & {\color{customred} 0.3579}\\
        
        1 & 0 & 10 & \faCheckCircle & 0.3060 $\pm$ 0.0054 & 0.0348 $\pm$ 0.0047 & 0 & \underline{0.3114}\\

        0.5 & 0.5 & 10 & \faTimesCircle & {\color{customred} 0.3428 $\pm$ 0.0144} & 0.2971 $\pm$ 0.0240 & 0.1487 $\pm$ 0.1594 & {\color{customred} 0.3572}\\

        0 & 1 & 10 & \faTimesCircle & 0.3391 {\color{customred} $\pm$ 0.0149} & 0.2874 $\pm$ 0.0251 & 0.2177 $\pm$ 0.1027 & {\color{customred} 0.3540}\\
        0 & 1 & 10 & \faCheckCircle & 0.3025 $\pm$ 0.0067 & 0.2795 $\pm$ 0.0766 & 0 & \textbf{0.3092}\\
        
        \hline
        \multicolumn{8}{c}{Split ratio is fix at 0.7 and $\alpha$ is set to 0.34 ($\alpha_l$=0.3294).}\\
        \hline
        
        1 & 0 & 10 & \faTimesCircle & {\color{customred} 0.3404 $\pm$ 0.0168} & 0.2764 $\pm$ 0.0395 & 0.1212 {\color{customred} $\pm$ 0.2499} & {\color{customred} 0.3711}\\

        0.5 & 0.5 & 10 & \faTimesCircle & {\color{customred} 0.3407 $\pm$ 0.0157} & 0.2955 $\pm$ 0.0378 & 0.1350 {\color{customred} $\pm$ 0.2069} & {\color{customred} 0.3564}\\

        0.5 & 0.5 & 20 & \faTimesCircle & {\color{customred} 0.3402 $\pm$ 0.0102} & 0.2916 $\pm$ 0.0337 & 0.1160 $\pm$ 0.1713 & {\color{customred} 0.3504}\\

        0.5 & 0.5 & 20 & \faCheckCircle & 0.3023 $\pm$ 0.0112 & 0.2665 $\pm$ 0.0293 & 0 & \underline{0.3135}\\

        0 & 1 & 20 & \faTimesCircle & 0.3382 {\color{customred} $\pm$ 0.0154} & 0.2927 $\pm$ 0.0353 & 0.1287 $\pm$ 0.1156 & {\color{customred} 0.3536}\\

        0 & 1 & 20 & \faCheckCircle & 0.3006 $\pm$ 0.0121 & 0.2539 $\pm$ 0.0109 & 0 & \textbf{0.3127}\\
    
        \bottomrule
    \end{tabular}
}
\end{table*}

%% file: Tables/Vicuna_conditional.tex
\begin{table}[t!]
\centering
\caption{Mean of SSM results obtained from 100 trials at the risk level of 0.3 on the Clinical Knowledge subject of MMLU dataset. Note that we fix the split ratio to 0.5 and set $w_l=w_f=0.5$ in the formulation of NS.}\label{table: vicuna7_13 conditional coverage}
\adjustbox{max width=\linewidth}{
    \begin{tabular}{ccccc} 
        \toprule
        \textbf{LLMs} & \textbf{OD} & \textbf{Size $=$ 1} & \textbf{Size $=$ 2} & \textbf{Size} $=$ \textbf{3}\\
        \midrule

        Vicuna-7B-v1.5 & \faTimesCircle & 0.3233 & 0.3811 & 0.2113\\
        ($a_l=0.2857$) & \faCheckCircle & 0.3229 & 0.2971 & 0.0733\\
        
        Vicuna-13B-v1.5 & \faTimesCircle & 0.3045 & 0.2769 & 0.2811 \\
        ($a_l=0.2556$) & \faCheckCircle & \textbf{0.2973} & \textbf{0.1813} & \textbf{0}\\
        
        \bottomrule
    \end{tabular}
}
\end{table}

%% file: Tables/Semantic_redundancy.tex
\begin{table}[t!]
\centering
\caption{Results of mean APSS before and after semantic deduplication (SD) within prediction sets.}\label{table: semantic redundancy}
\adjustbox{max width=\linewidth}{
    \begin{tabular}{cccc} 
        \toprule
        \textbf{Dataset} & \textbf{LLMs} & \textbf{SD} & \textbf{Mean APSS}\\
        \midrule

        \multirow{6}{*}{TriviaQA} & \multirow{2}{*}{Qwen2.5-3B-Instruct} & \faTimesCircleO & 8.07\\
        {} & {} & \faCheckCircleO & 1.08\\

        {} & \multirow{2}{*}{Qwen2.5-7B-Instruct} & \faTimesCircleO & 8.77\\
        {} & {} & \faCheckCircleO & 1.05\\
        
        {} & \multirow{2}{*}{Qwen2.5-14B-Instruct} & \faTimesCircleO & 9.18\\
        {} & {} & \faCheckCircleO & 1.03\\

        \hline
        
        \multirow{6}{*}{CoQA} & \multirow{2}{*}{Qwen2.5-3B-Instruct} & \faTimesCircleO & 8.65\\
        {} & {} & \faCheckCircleO & 1.12\\

        {} & \multirow{2}{*}{Qwen2.5-7B-Instruct} & \faTimesCircleO & 8.80\\
        {} & {} & \faCheckCircleO & 1.04\\
        
        {} & \multirow{2}{*}{Qwen2.5-14B-Instruct} & \faTimesCircleO & 8.84\\
        {} & {} & \faCheckCircleO & 1.03\\
    
        \bottomrule
    \end{tabular}
}
\end{table}

%% file: Sections/Conclusion.tex
\section{Conclusion}
\label{sec: Conclusion}
In this paper, we introduce SConU, a modular and principled framework aimed at eliminating uncertainty data outliers that violate the exchangeability precondition inherent in existing conformal approaches. 
We develop two conformal p-values to identify whether the given test QA sample significantly deviates from the uncertainty distribution of the calibration set as a user-specified risk level. 
Experimental results demonstrate the rigorous guarantees of marginal coverage and efficient prediction of SConU. 
Additionally, we derive the minimum risk level manageable by the calibration set without manually handling calibration data points post-deployment of the language model. 
Furthermore, we approximate conditional coverage across various sizes of the prediction set by analyzing several internal components of the conformal procedures.

%% file: Sections/Appendix.tex
\section{An illustration of the SCP framework}\label{sec: SCP in classification}
SCP can transform any heuristic notion of uncertainty from any model into a rigorous one~\citep{angelopoulos2021gentle}. 
Let’s illustrate the basic SCP framework by classification problems~\citep{angelopoulos2021uncertainty}: Given the calibration set of size $N$, we define the NS of each sample as one minus the softmax output for the true class. 
Then we calculate the $\frac{\left \lceil \left ( N+1\right )\left ( 1-\alpha \right )\right \rceil}{N}$ quantile of the $N$ sorted (ascending) NSs and employ it as the threshold to select possible classes for a new test sample. 
If the softmax output of a certain class falls below the threshold, by the exchangeability condition, we consider it to have an approximate probability of $1-\alpha$ to be the true label and add it to the prediction set. 
Finally, we achieve marginal correctness coverage on the finite-sample test set. 
The complete framework is presented as follows:
\begin{enumerate}
    \item Given the calibration data set ${\left \{\left ( X_i, {Y}_{i}^{*}\right ) \right \}}_{i=1}^{n}$ (i.i.d.) and pretrained model $\hat{f}\left ( \cdot\right )$ ($\hat{f}\left ( X_i\right ) \in {\left [ 0, 1\right ]}^{\left ( K\right )}$). 
    The probability of each true class (label) is denoted as ${\hat{f}\left ( X_i\right ) }_{{Y}_{i}^{*}}$. 
    \item Define and sort the nonconformity scores (uncertainty state associated with the true class of each calibration sample): $s_i= s\left ( X_i, {Y}_{i}^{*}\right ) =1-{\hat{f}\left ( X_i\right ) }_{{Y}_{i}^{*}}$ ($\left \{s_1 \leq \cdots \leq s_n \right \}$).
    \item Obtain the $\frac{\left \lceil \left ( n+1\right )\left ( 1-\alpha \right )\right \rceil}{n}$ quantile of ${\left \{ s_i\right \}}_{i=1}^{n}$: $\hat{q}=\inf \left \{ q:\frac{\left | \left \{ i:s_i \leq  q\right \}\right |}{n} \geq \frac{\left \lceil \left ( n+1\right )\left ( 1-\alpha \right )\right \rceil}{n} \right \}=$\\
    ${s}_{\left \lceil \left ( n+1\right )\left ( 1-\alpha \right )\right \rceil}$.
    \item Create the prediction set for ${X}_{test}$ following: $\mathcal{C}\left ( {X}_{test}\right )=\left \{ y\in[K]: s\left ( {X}_{test}, y\right )\leq \hat{q}\right \}$
    \item The event ${Y}_{test}^{*} \in \mathcal{C}\left ( {X}_{test}\right ) $ is equivalent to $ s\left ( {X}_{test}, {Y}_{test}^{*}\right )\leq \hat{q}$. 
    As long as $s\left ( {X}_{test}, {Y}_{test}^{*}\right )\leq \hat{q}$ is satisfied, ${Y}_{test}^{*}$ is encompassed by $\mathcal{C}\left ( {X}_{test}\right )$, and then we obtain the prediction set that contains the true label. 
    \item By the exchangeability of $N+1$ data points, we have $\mathbb{P}\left ( {s}_{test} \leq s_i \right )=\frac{i}{n+1}$.
    \item Then we conclude:  $\mathbb{P}\left ( {Y}_{test}^{*} \in \mathcal{C}\left ( {X}_{test}\right ) \right )=\mathbb{P}\left ( {s}_{test} \leq \hat{q} \right )=\frac{\left \lceil \left ( n+1\right )\left ( 1-\alpha \right )\right \rceil}{n+1}\geq 1-\alpha$. 
\end{enumerate}

\section{Additional Experimental Settings}\label{sec: Additional Experimental Settings}
\subsection{Base LLMs}
We conduct experiments utilizing 4 popular series of ``off-the-shelf'' LLMs: OpenChat~\citep{wang2024openchat}, LLaMA~\citep{Touvron2023Llama2O,llama3modelcard}, Vicuna~\citep{vicuna2023}, and Qwen~\citep{Yang2024Qwen2TR}, divided by model size into: \ding{172} 3B: LLaMA-3.2-3B-Instruct and Qwen-2.5-3B-Instruct. \ding{173} 7B: Qwen-2-7B-Instruct, Qwen-2.5-7B-Instruct, and OpenChat-3.5. \ding{174} 8B: LLaMA-3-8B-Instruct and LLaMA-3.1-8B-Instruct. \ding{175} 13B: LLaMA-2-13B-Chat and Vicuna-13B-v1.5. \ding{176} 14B: Qwen-2.5-14B-Instruct. \ding{177} 32B: Qwen-2.5-32B-Instruct. 

\subsection{Details of Datasets}
\label{sec: Details of Datasets}
\textbf{MMLU}\footnote{\href{https://huggingface.co/datasets/cais/mmlu}{https://huggingface.co/datasets/cais/mmlu}} is a massive multi-task test consisting of multiple-choice questions from  57 subjects such as anatomy, astronomy, and business ethics. 
Following prior studies~\cite{kumar2023conformal, su-etal-2024-api}, we consider a subset of 16 subjects: computer security, high school computer science, college computer science, machine learning, formal logic, high school biology, anatomy, clinical knowledge, college medicine, professional medicine, college chemistry, marketing, public relations, management, business ethics, and professional accounting. 
Table~\ref{table: mmlu dataset} presents the number of samples employed for each subject from the MMLU dataset. 
Note that there is a slight deviation in the actual number of samples utilized for each model due to a few individual samples that do not comply with user instructions in all sampled responses (i.e., each response is not among A, B, C, or D).

\input{Tables/MMLU}

\textbf{MMLU-Pro}\footnote{\href{https://huggingface.co/datasets/TIGER-Lab/MMLU-Pro}{https://huggingface.co/datasets/TIGER-Lab/MMLU-Pro}} is a more robust and challenging multi-task understanding dataset. 
It expands samples from MMLU by increasing the 4 options for each question to 10, and the subjects are enhanced with questions from STEM Website, TheoremQA, and SciBench. 
This dataset totally contains 12,000 complex questions across various disciplines. 
qIn order for a balanced distribution of sample quantities across different subjects, we employ a maximum of 500 samples for each subject. 
The detailed sample quantities are shown in Table~\ref{table: mmlu-pro dataset}. 
Note that the number of samples applied for each model may have slight deviations (i.e., each response is not among A, B, C, D, E, F, G, H, I, or J).

\input{Tables/MMLU_Pro}

For both MMLU and MMLU-Pro datasets, we utilize the test set of each subject, sourced from the \texttt{test-00000-of-00001.parquet} file. 

\textbf{MedMCQA}\footnote{\href{https://huggingface.co/datasets/openlifescienceai/medmcqa}{https://huggingface.co/datasets/openlifescienceai/medmcqa}} is designed to address real-world medical entrance exam questions. 
We consider the full validation set, 4,180 MCQA samples, sourced from the \texttt{validation-00000-of-00001.parquet} file. 
Note that several MCQA samples cannot be correctly encoded by the \texttt{tokenizer}, specifically non-ASCII characters. 
We exclude these samples, remaining 3,967 samples. 

\textbf{TriviaQA}\footnote{\href{https://huggingface.co/datasets/mandarjoshi/trivia_qa}{https://huggingface.co/datasets/mandarjoshi/triviaqa}} is a reading comprehension dataset containing over 650,000 high-quality query-answer pairs. 
We utilize the validation set sourced from the \texttt{validation-00000-of-00001.parquet} file and randomly select 4,000 QA samples. 
In the experiments of sampling size calibration, we only employ 2,000 samples.

\textbf{CoQA}\footnote{\href{https://huggingface.co/datasets/stanfordnlp/coqa}{https://huggingface.co/datasets/stanfordnlp/coqa}} is a large-scale conversational QA task, including 127,000 query-answer samples with their corresponding evidence highlighted in the provided context. 
We also utilize the validation set sourced from the \texttt{validation-00000-of-00001.parquet} file and randomly select 4,000 QA samples.

\subsection{Prompt Engineering}
\label{sec: Prompt Engineering}
For both the MMLU and MMLU-Pro tasks, we randomly select 3 QA examples from the validation set of each subject, to construct a 3-shot prompt, which guides the language model in answering the current question using the specified response format (i.e., providing options like A, B, or C). 
Notably, all questions within the same subject share the same examples in the 3-shot prompt. 
For the MedMCQA task, we randomly selected 3 samples from the validation set as few-shot examples and exclude these three samples from subsequent experiments. 
We apply similar system prompts across the 3 MCQA datasets. 
Note that each question in the MMLU-Pro dataset generally includes 10 multiple-choice options, though some QA samples have fewer options following a manual review process to eliminate unreasonable choices. 
In the TriviaQA and CoQA tasks, we develop few-shot prompts following prior work~\citep{duan-etal-2024-shifting,WANG2025109553}. 
We provide complete prompt examples for 5 datasets, as presented in Figures~\ref{fig: mmlu prompt example}$-$\ref{fig: CoQA prompt example}.

\subsection{Hyperparameters}
Following prior studies~\cite{duan-etal-2024-shifting, wang-etal-2024-conu, wang2024sample}, We employ multinominal sampling to generate $M$ candidate responses for each data point. 
For both the MMLU and MedMCQA datasets with 4 options for each question, we set the number of candidate responses, $M$, to 20, maintaining consistency with previous research~\cite{kuhn2023semantic,lin2024generating,quach2024conformal}. 
Since each sample in the MMLU-Pro dataset includes 10 multiple-choice options, we increase $M$ to 50 to better approximate the distribution of model outputs. 
For the TriviaQA and CoQA tasks, we generate 10 responses for each question~\citep{wang-etal-2024-conu}. 
Considering that we develop prompts to guide the language model in responding with the most probable option letters (e.g., A, B, or C), as detailed in Appendix~\ref{sec: Prompt Engineering}, we set the maximum generation length to 1 to accelerate sampling in 3 MCQA tasks. 
For open-domain QA, we examine the maximum length of answers for all randomly selected samples, and set the maximum generation length for 2 tasks to 36. In the \texttt{generate} function, we configure the hyperparameters as presented in Table~\ref{table: hyperparameters}. 
Moreover, since the conformal p-value detects when test points do not come from the same distribution of the calibration set, we guarantee that it does not return too many false positives and set $\delta$ equal to the user-specified risk level, following prior work~\citep{angelopoulos2021gentle,jin2023selection,gui2024conformal,Huang2024CONFINECP}. 

\input{Tables/Hyperparameters}

\section{Conformal p-value}
\label{sec: appendix Conformal p-value}
In this section, we first demonstrate that the conformal p-value formulated in Eq.~\eqref{eq: baseline_sconu} adheres to the statistical definition of p-values. 
As mentioned, $u_i$ represents the uncertainty of the LLM addressing the $i$-th question. 
At this point, we can denotes $p_{N+1}$ as
\begin{equation}
    \begin{split}
        p_{N+1} &= \frac{1+\displaystyle\sum_{i=1}^{N}\textbf{1} \left \{ u_i \geq u_{N+1}\right \}}{N+1}\\
        &=\frac{1+k}{N+1}
    \end{split},
\end{equation}
where $1+k$ is the position of $u_{N+1}$ in the sorted (i.e., ascending) sequence of $N+1$ uncertainty scores, and we have 
\begin{equation}\label{eq: eq 8}
    \begin{split}
        \mathbb{P}\left(p_{N+1} \leq \delta\right)&=\mathbb{P}\left(\frac{1+k}{N+1} \leq \delta\right)\\
        &=\mathbb{P}\left(1+k \leq \left \lfloor \left ( N+1\right ) \delta \right \rfloor \right)
    \end{split}.
\end{equation}
Since we apply the consistent uncertainty measure for each QA sample, the $N+1$ uncertainty scores are exchangeable. 
Then, we obtain
\begin{equation}
    \begin{split}
        \mathbb{P}\left(p_{N+1} \leq \delta\right)&=\frac{\left \lfloor \left ( N+1\right ) \delta \right \rfloor}{N+1}\\
        &\leq \frac{\left(N+1\right)\delta}{N+1}\\
        &\leq\delta
    \end{split}.
\end{equation}

As mentioned in section~\ref{sec: Empirical Results}, we observe minor fluctuations in the results of SConU under cross-domain scenarios. 
This arises from the hallucination issues of LLMs. 
For example, consider two questions with similar sampling distribution. 
However, in one question's candidate set, nearly all the answers are incorrect, while in the other question's sampling set, most answers are correct. 
In this case, the scores obtained from the uncertainty measure for the two samples may be the same, but in fact, the answering situations of the two QA samples are opposite, which can affect the exchangeability of the uncertainty scores, leading to slight variations in the performance of outlier detection. 

To check whether the uncertainty score of each calibration data point is referenceable at different risk levels, we incorporate their prediction status into the counting criterion. 
At this point, we denote the count of calibration samples that satisfy both $u_i\geq u_{N+1}$ and $y_i^* \in E\left( x_i, \mathcal{D}_{cal}, \alpha \right)$. 
Thus the conformal p-value can be expressed as $p^{'}_{N+1}=\frac{1+k}{N+1}$. 
Here, $k$ can take values from $0$ to $N$, so the range of $p^{'}_{N+1}$ is $\left[\frac{1}{N+1},1\right]$. 
Similar to Eq.~\eqref{eq: eq 8}, we have 
\begin{equation}
        \mathbb{P}\left(\frac{1+k}{N+1} \leq \delta\right)=\mathbb{P}\left(k \leq  \left ( N+1\right ) \delta-1  \right).
\end{equation}
Let $m=\left \lfloor \left ( N+1\right ) \delta-1 \right \rfloor$. 
Since $k$ can be at most $N$, if $M<0$, then $p_{N+1}$ will always be greater than any negative value, so $\mathbb{P}\left(p^{'}_{N+1}\leq\delta\right)=0\leq \delta$. 
If $0\leq m \leq N$, we have 
\begin{equation}
    \mathbb{P}\left(k \leq m \right) \leq \frac{m+1}{N+1}.
\end{equation}
Therefore, 
\begin{equation}
    \begin{split}
        \mathbb{P}\left(p^{'}_{N+1}\leq \delta\right) &\leq \frac{m+1}{N+1}\\
        & \leq \delta
    \end{split}.
\end{equation}
In summary, we have demonstrated that our developed two conformal p-values satisfy the statistical definition of p-values. 

\input{Tables/mmlu_risk_level}

\begin{figure*}[!t]
    \centering
     \begin{subfigure}{0.495\linewidth}
        \centering
        \includegraphics[width=\linewidth]{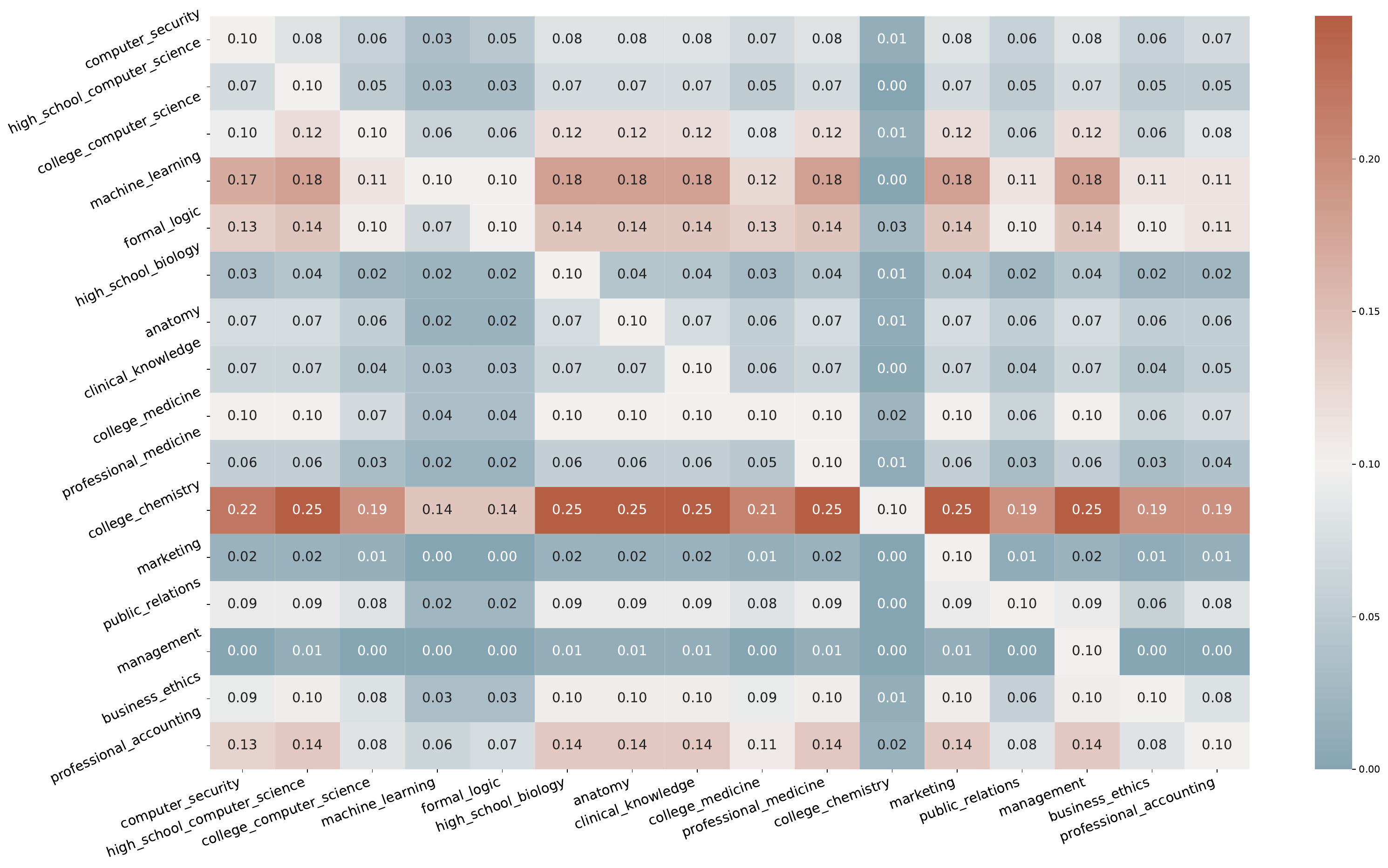}
        \caption{EMR results of the basic ConU framework.}
	\label{fig: cross-domain original mmlu emr}
    \end{subfigure}
    \hfill
    \centering
    \begin{subfigure}{0.495\linewidth}
	\centering
	\includegraphics[width=\linewidth]{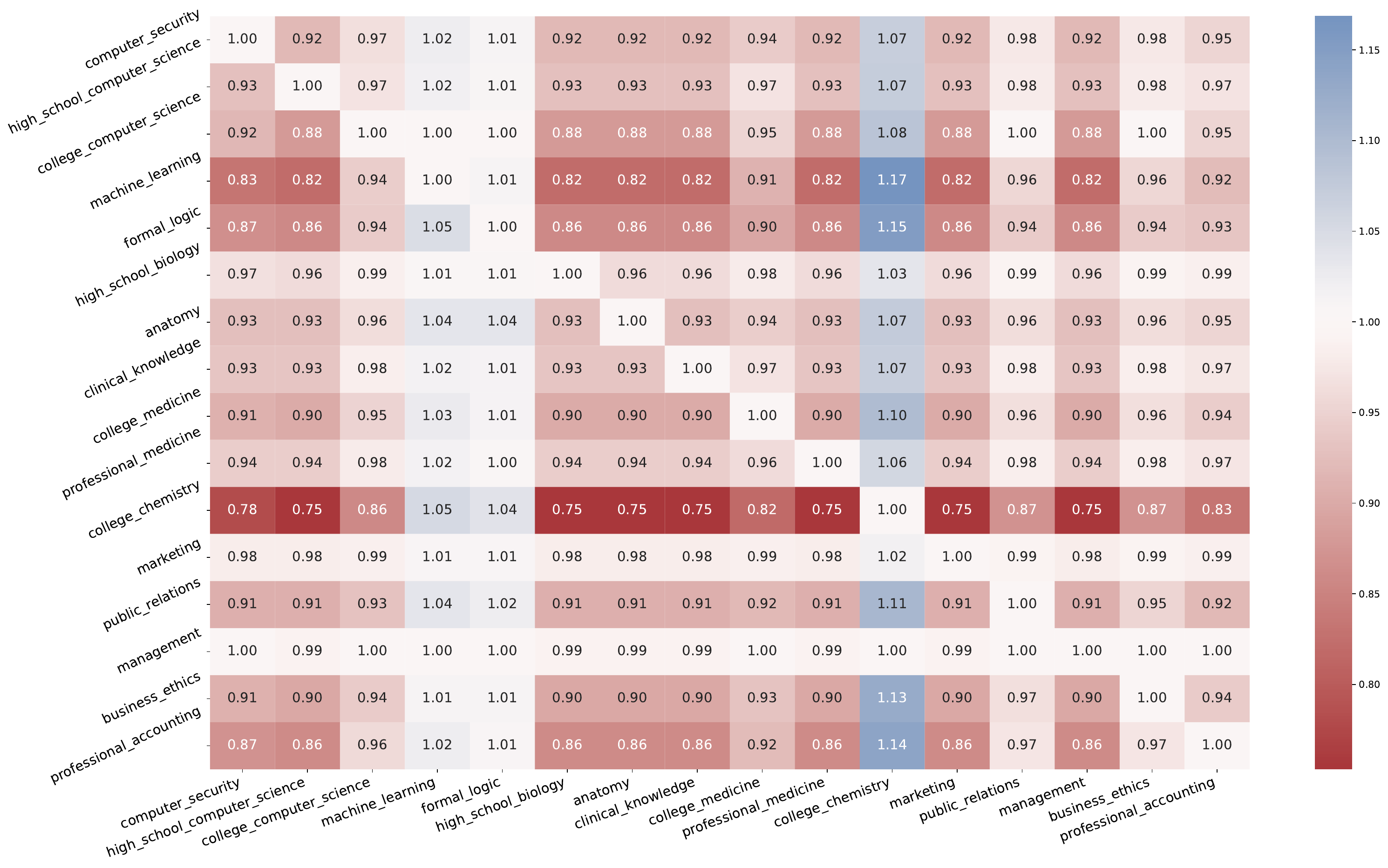}
	\caption{APSS results of the basic ConU framework.}
	\label{fig: cross-domain original mmlu apss}
    \end{subfigure}
    \caption{Results of the EMR and APSS metrics obtained from the basic ConU framework on the MMLU dataset utilizing the Qwen2.5-32B-Instruct model at the risk level of 0.1.}\label{fig: cross-domain mmlu original}
\end{figure*}

\begin{figure*}[!t]
    \centering
     \begin{subfigure}{0.495\linewidth}
        \centering
        \includegraphics[width=\linewidth]{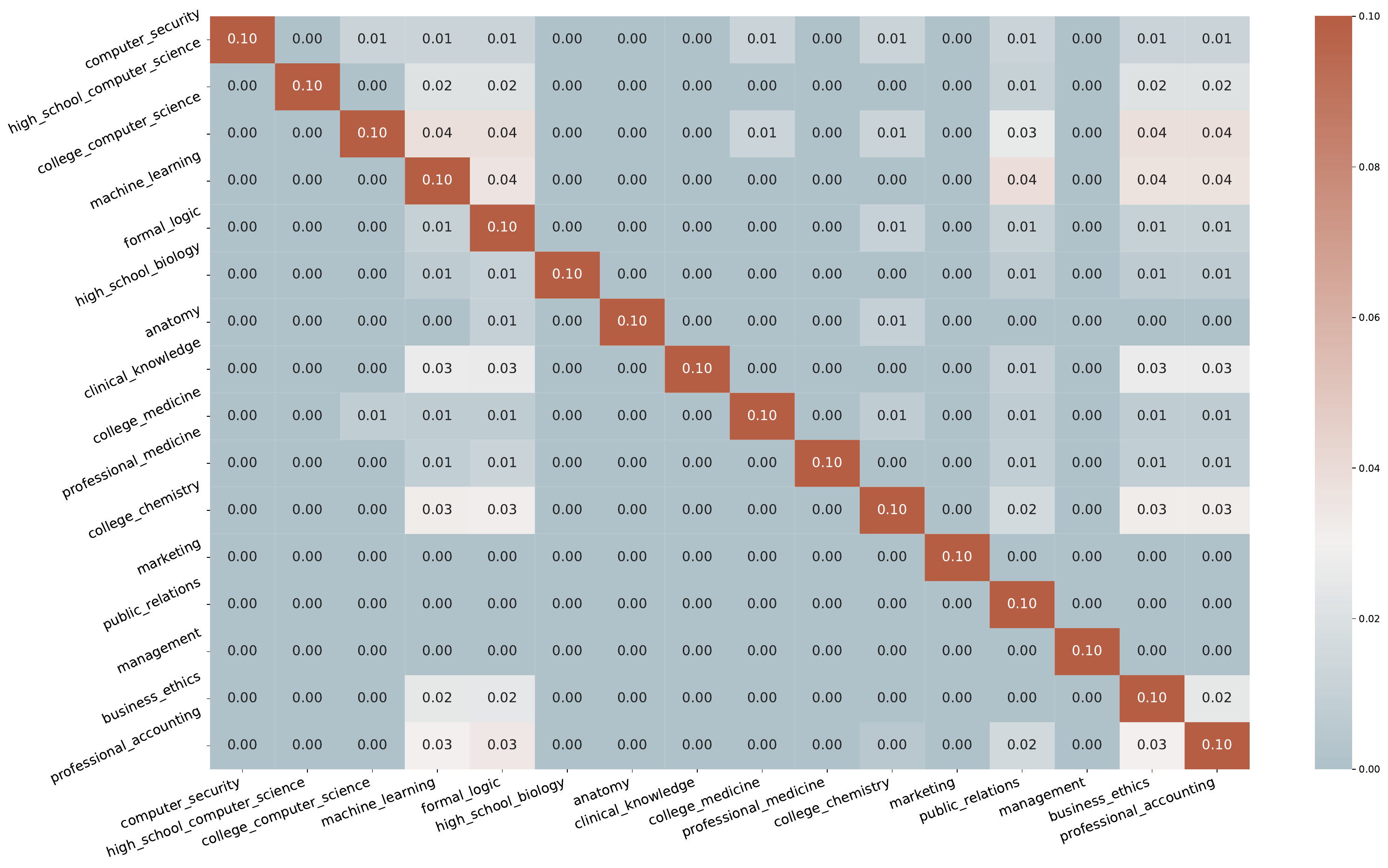}
        \caption{EMR results of our SConU framework.}
	\label{fig: cross-domain sconu mmlu emr}
    \end{subfigure}
    \hfill
    \centering
    \begin{subfigure}{0.495\linewidth}
	\centering
	\includegraphics[width=\linewidth]{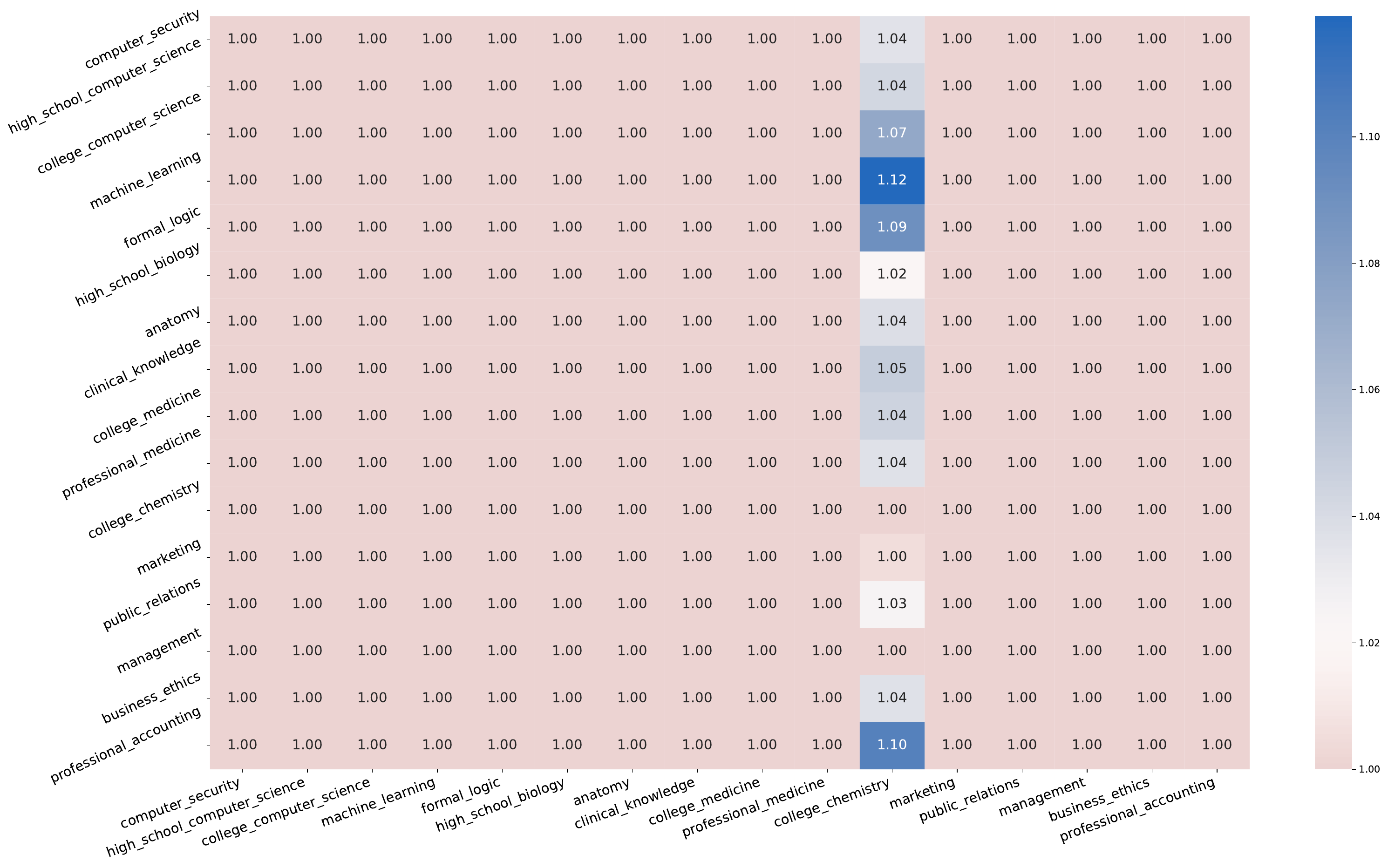}
	\caption{APSS results of our SConU framework.}
	\label{fig: cross-domain sconu mmlu apss}
    \end{subfigure}
    \caption{Results of the EMR and APSS metrics obtained from our SConU framework on the MMLU dataset utilizing the Qwen2.5-32B-Instruct model at the risk level of 0.1.}\label{fig: cross-domain mmlu sconu}
\end{figure*}

Considering that a single uncertainty notion cannot fully represent the exchangeability among QA samples, we can perform multiple hypothesis testing to identify uncertainty data outliers in practical high-stakes QA applications. 
As mentioned in Section~\ref{sec: Selective Conformal Uncertainty}, we utilize PE as the uncertainty measure, formulated as $u_i=PE\left( x_i \right)=\textstyle\sum_{o=1}^{O_i}-p_o \log p_o$, where $p_o$ denotes the logit-based confidence score of the $o$-th option and $O_i$ denotes the number of options for the $i$-th question (e.g., 4 or 10). 
Here, for each QA sample, we define $B$ notions of uncertainty: $\left\{u^{(i)}_b\right\}_{b=1}^B$, such as the number of semantics within the candidate set~\citep{lin2024generating} and the frequency-based PE. 
At this point, we check whether its $B$ types of uncertainty significantly deviate from the calibration set for each test data point. If any one of them does not meet the criterion, we consider that the exchangeability condition is violated and decline to provide a prediction set. 

We determine the significance level for the p-value associated with each uncertainty notion by the  Benjamini-Hochberg (BH) procedure~\citep{benjamini1995controlling,benjamini2001control}. 
More details can be referred to the study~\citep{jin2023selection}. 
Finally, for each test QA sample, if a certain conformal p-value associated with one uncertainty notion is lower than the significance level calculated by the BH procedure, we reject the null hypothesis and decline to provide an answer. 
Conversely, when multiple hypothesis testing indicates that the $N+1$ QA samples are exchangeable, we select task-specific ConU methods. 
Next, we present several typical frameworks.

\section{Details of Conformal Procedures}\label{sec: Details of Utilizing the Conformal Frameworks}
Similar to Prompt Risk Control (PRC)~\cite{zollo2024prompt}, our approach is orthogonal to some existing conformal frameworks. 
For MCQA tasks within the same discipline or dataset, we apply the basic procedures in prior studies~\citep{kumar2023conformal,ye2024benchmarking,kostumov2024uncertainty} and evaluate the EMR metric before and after implementing our developed conformal p-value. 
In formulation, the NS of each option can be expressed as $1 - w_l \cdot F_l \left(y_i^*\right) - w_f \cdot F_f \left(y_i^*\right)$ as defined in Section~\ref{sec: Preliminaries}. 
Here, we only utilize the confidence score obtained from the model logit and set $w_l=1$ and $w_f=0$. 
At this point, we calculate the uncertainty score based on the logit-based PE method.

In more practical cross-domain settings, we investigate employing the black-box frequency score to formulate the NS following the research~\citep{wang2024sample}, assuming no access to model internal information, and set $w_l=0$ and $w_f=1$. 
We use the frequency score of each option obtained from the candidate set of size 20 (or 50) to characterize the probability of $p_o$ and calculate frequency-based PE to implement uncertainty data outlier detection. 
Note that the performance of uncertainty quantification methods to differentiate between correct and incorrect answers affects the effectiveness of conformal p-values in identifying outliers. 
This is because a single notion of uncertainty cannot fully characterize the exchangeability among data points. 
Therefore, by applying more efficient uncertainty measures~\citep{lin2024generating,duan-etal-2024-shifting,WANG2025109553}, we can enhance the capability of the NS to represent the disagreement between the current question and response while also improving the statistical rigor of significance tests.

In open-domain QA tasks, we employ the similar ConU framework applicable to black-box settings introduced in the study~\citep{wang-etal-2024-conu}. 
The NS of each calibration data is formulated as $1-0.5 \cdot F \left( y_{ref}^{(i)} \right) -0.5 \cdot \frac{1}{M}\textstyle\sum_{j=1}^{M} S\left(y_{ref}^{(i)}, y_j^{(i)}\right)F\left(y_j^{(i)}\right)$, where $y_{ref}^{(i)}$ represent the response in the candidate set that have equivalent semantics to the ground-truth $y_i^*$, $F \left( y_{ref}^{(i)} \right)$ measures the number of generations that is semantically equivalent to $y_{ref}^{(i)}$ (i.e., the frequency score of correct semantic), and $S\left(y_{ref}^{(i)}, y_j^{(i)}\right)$ measures the semantic similarity score between $y_{ref}^{(i)}$ and $y_j^{(i)}$ in the candidate set. 
Refer to the studies~\citep{wang-etal-2024-conu,wang2024sample,su-etal-2024-api} for more details. 
We also link the NS with the uncertainty state of acceptable semantics. 

\input{Tables/mmlu_pro_risk_level}

\section{Additional Experimental Results}\label{sec: additional result in appendix}

\begin{figure*}[!t]
    \centering
     \begin{subfigure}{0.495\linewidth}
        \centering
        \includegraphics[width=\linewidth]{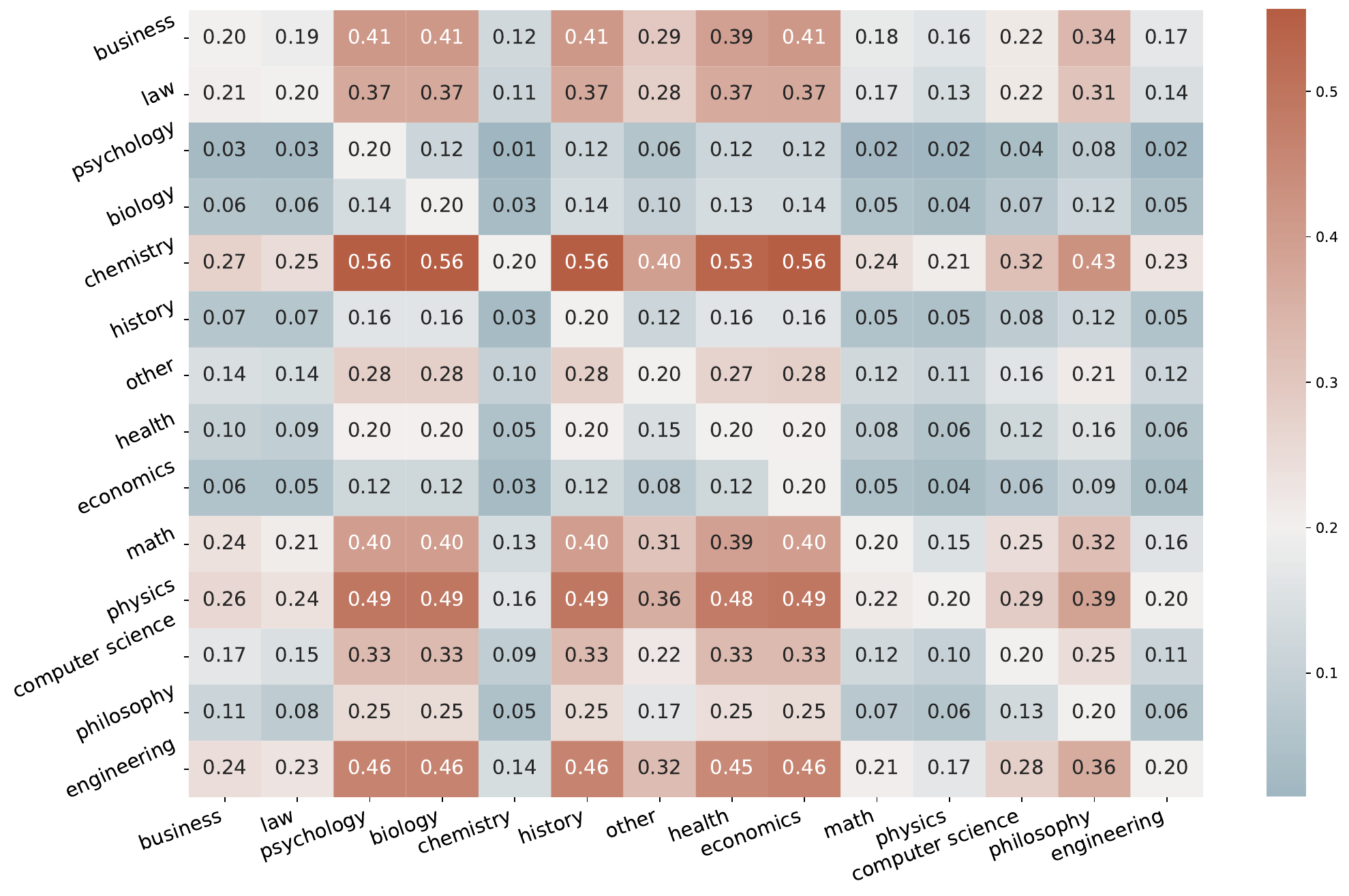}
        \caption{EMR results of the basic ConU framework.}
	\label{fig: cross-domain original mmlu-pro emr 32B}
    \end{subfigure}
    \hfill
    \centering
    \begin{subfigure}{0.495\linewidth}
	\centering
	\includegraphics[width=\linewidth]{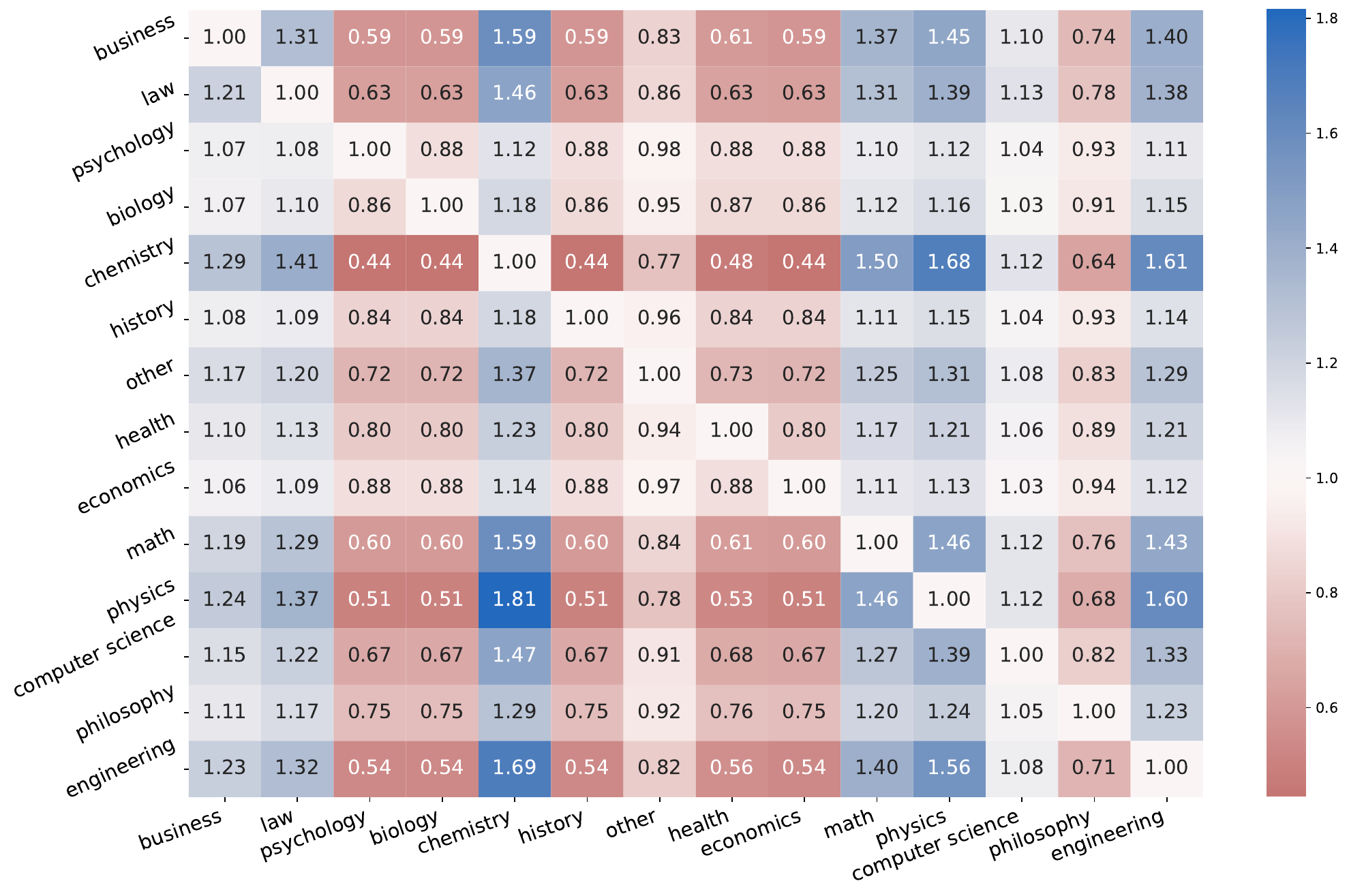}
	\caption{APSS results of the basic ConU framework.}
	\label{fig: cross-domain original mmlu-pro apss 32B}
    \end{subfigure}
    \caption{Results of the EMR and APSS metrics obtained from the basic ConU framework on the MMLU-Pro dataset utilizing the Qwen2.5-32B-Instruct model at the risk level of 0.2.}\label{fig: cross-domain mmlu-pro original 32B}
\end{figure*}

\begin{figure*}[!t]
    \centering
     \begin{subfigure}{0.495\linewidth}
        \centering
        \includegraphics[width=\linewidth]{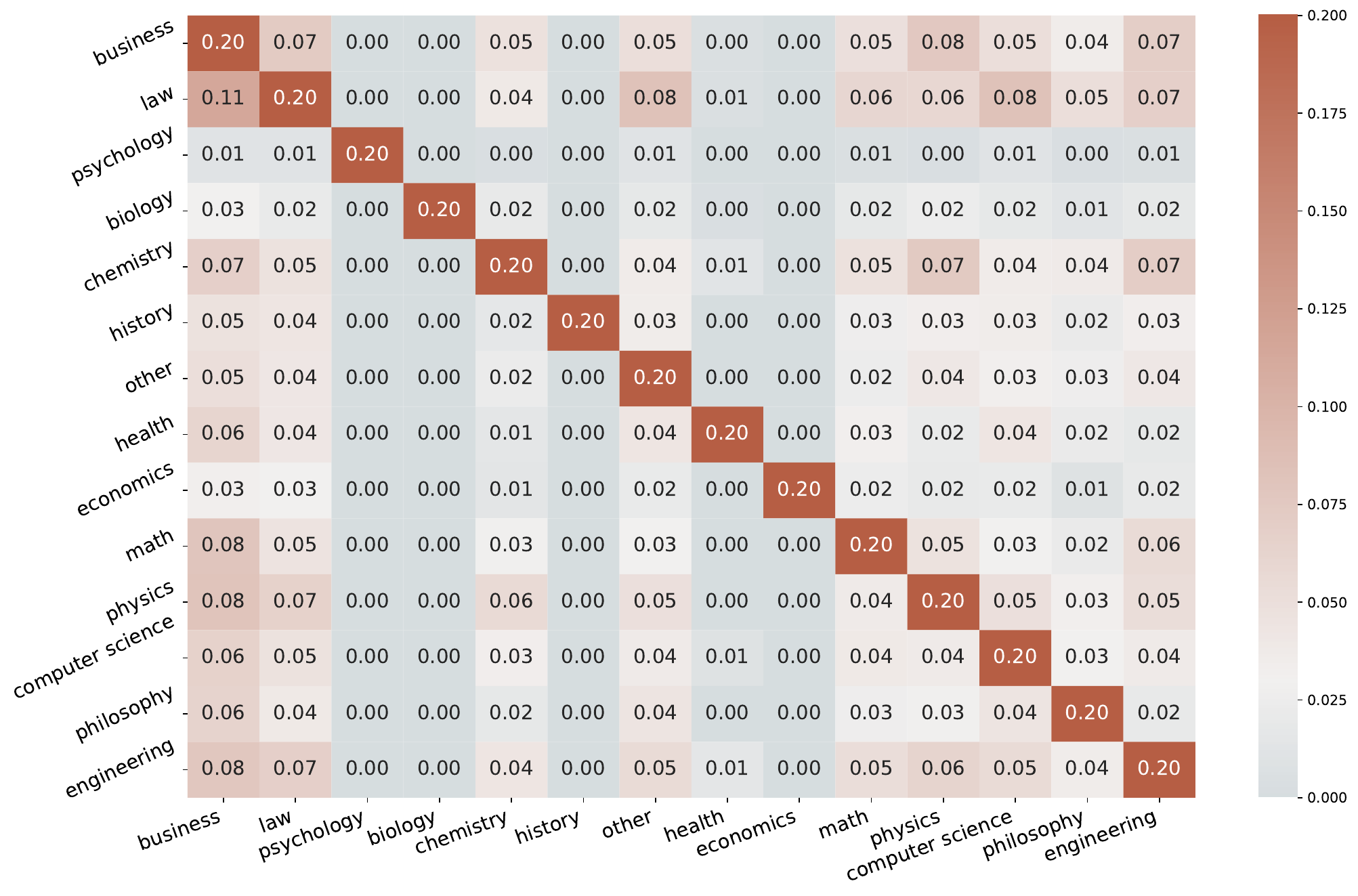}
        \caption{EMR results of our SConU framework.}
	\label{fig: cross-domain sconu mmlu-pro emr 32B}
    \end{subfigure}
    \hfill
    \centering
    \begin{subfigure}{0.495\linewidth}
	\centering
	\includegraphics[width=\linewidth]{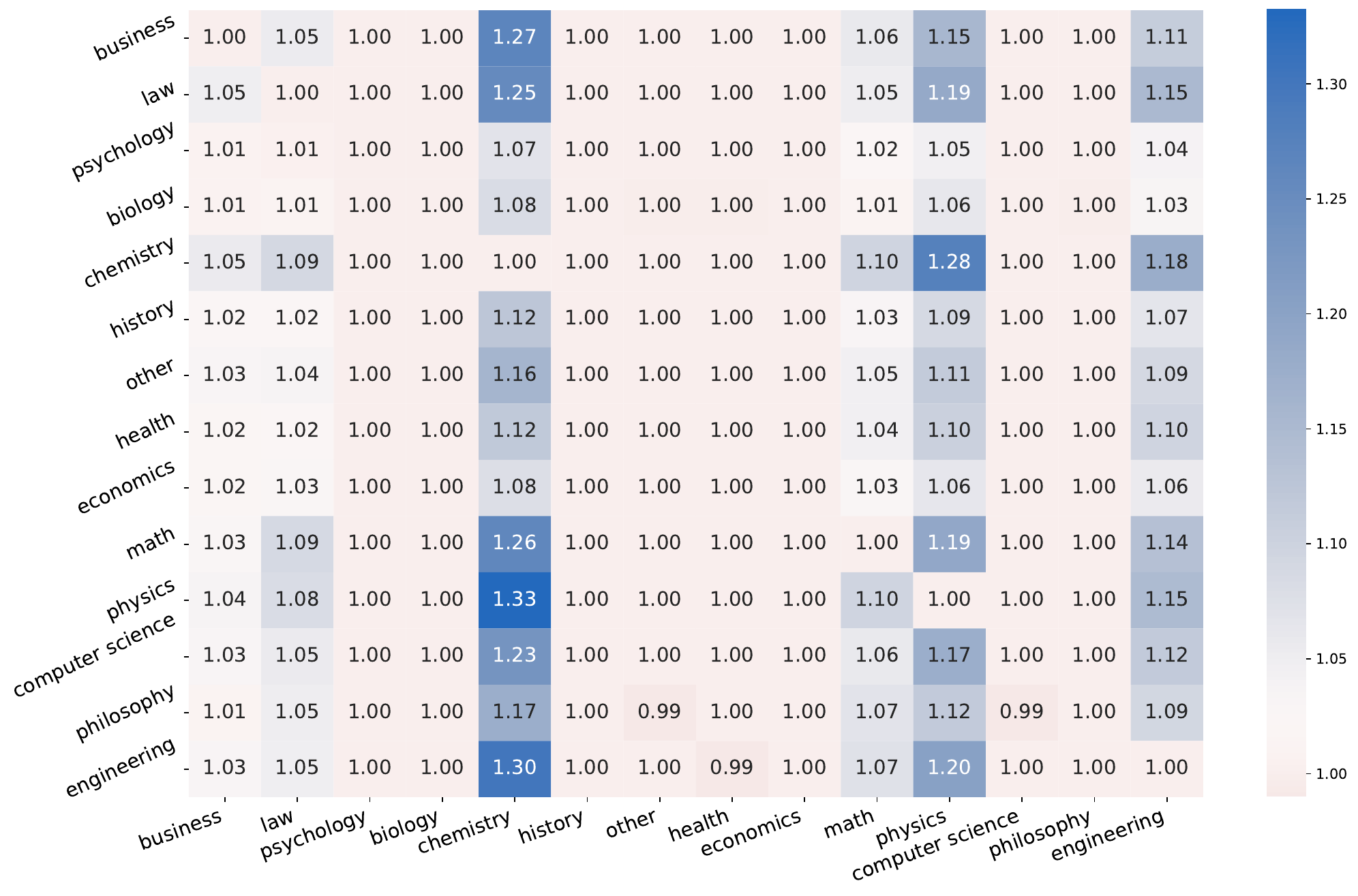}
	\caption{APSS results of our SConU framework.}
	\label{fig: cross-domain sconu mmlu-pro apss 32B}
    \end{subfigure}
    \caption{Results of the EMR and APSS metrics obtained from our SConU framework on the MMLU-Pro dataset utilizing the Qwen2.5-32B-Instruct model at the risk level of 0.2.}\label{fig: cross-domain mmlu-pro sconu 32B}
\end{figure*}

\begin{figure*}[!t]
    \centering
     \begin{subfigure}{0.495\linewidth}
        \centering
        \includegraphics[width=\linewidth]{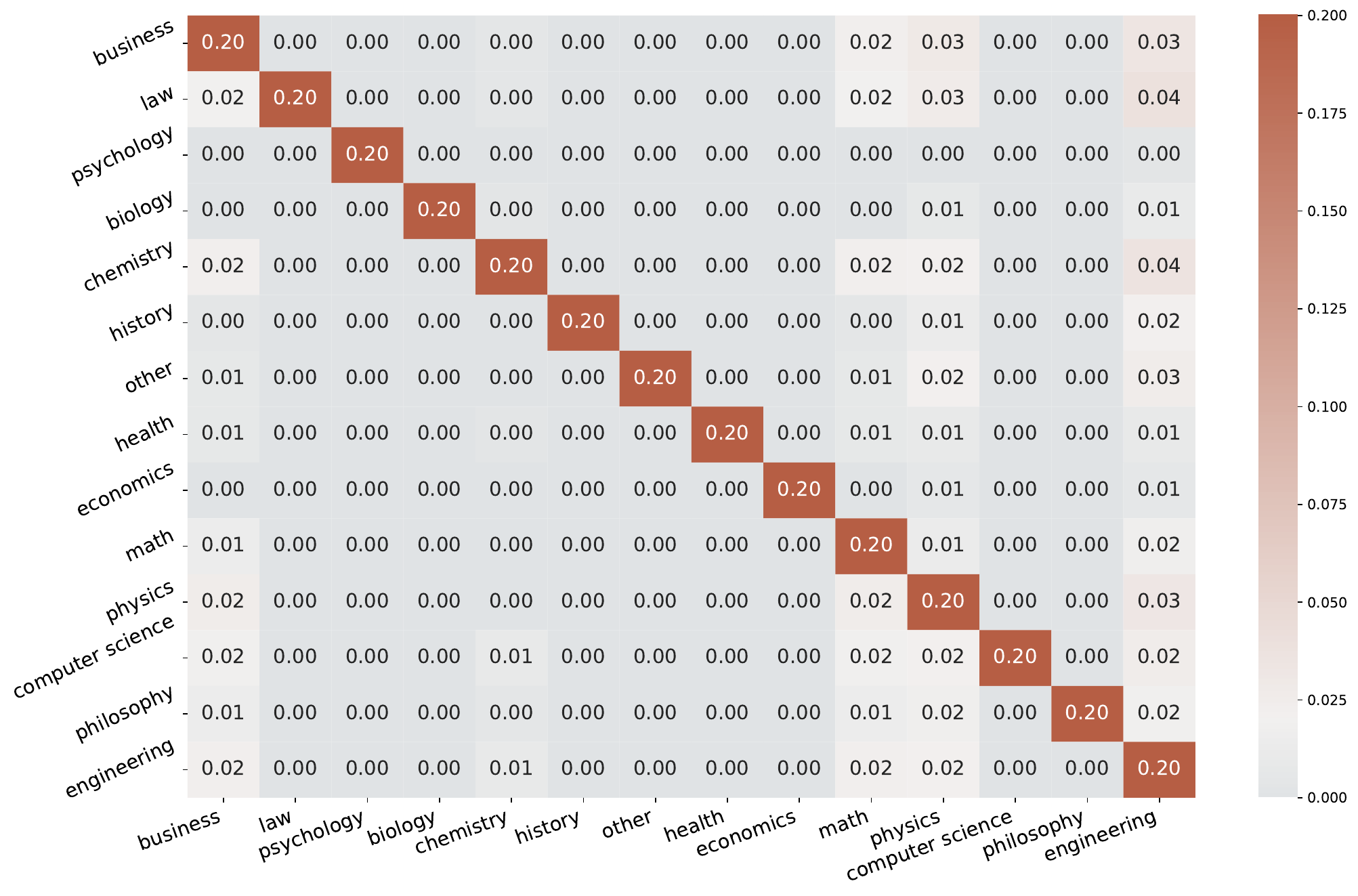}
        \caption{EMR results of our SConU-Pro framework.}
	\label{fig: cross-domain sconu-pro mmlu-pro emr 32B}
    \end{subfigure}
    \hfill
    \centering
    \begin{subfigure}{0.495\linewidth}
	\centering
	\includegraphics[width=\linewidth]{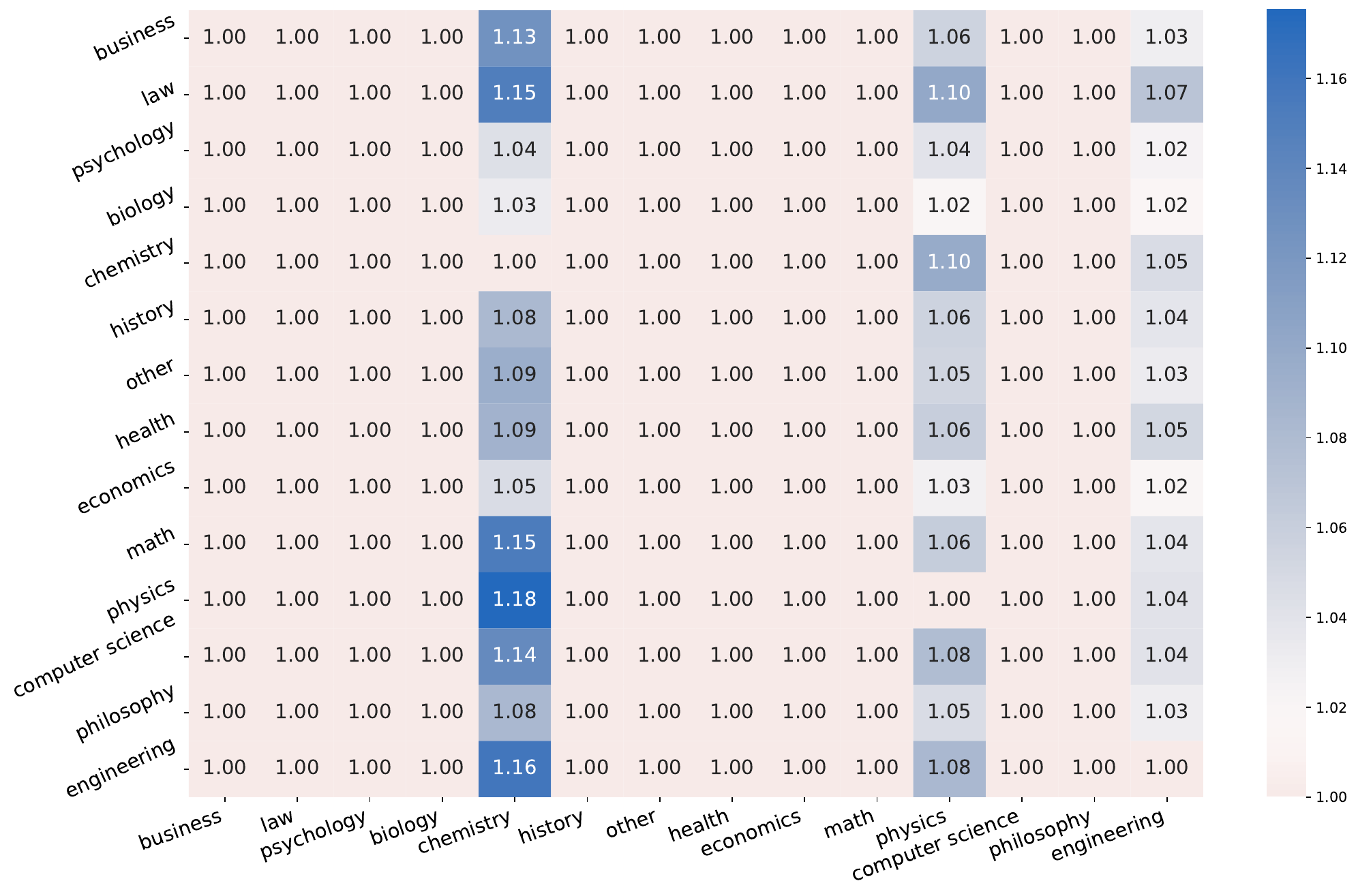}
	\caption{APSS results of our SConU-Pro framework.}
	\label{fig: cross-domain sconu-pro mmlu-pro apss 32B}
    \end{subfigure}
    \caption{Results of the EMR and APSS metrics obtained from our SConU-Pro framework on the MMLU-Pro dataset utilizing the Qwen2.5-32B-Instruct model at the risk level of 0.2.}\label{fig: cross-domain mmlu-pro sconu-pro 32B}
\end{figure*}

Note that in all experimental results within the interdisciplinary scenarios, the discipline from the horizontal axis represents the calibration set, while the discipline from the vertical axis represents the test set. 
When the calibration set and test set belong to the same discipline along the diagonal, all EMR results are directly set equal to the corresponding risk level of $\alpha$, and the APSS results are set to 1. 

In this section, we also evaluate our SConU framework in cross-domain settings utilizing the MMLU dataset with the Qwen2.5-32B-Instruct model employed as the generator. 
Firstly, we calculate the minimum risk level manageable by each subject of the calibration set based on Eq.~\eqref{eq: minimum risk level}, as presented in Table~\ref{table: mmlu minimum risk level}. 
Then, we specify the risk level to 0.1 ($\alpha=\delta=0.1$) and evaluate the results of the EMR metric on each subject of the test set before and after performing our SConU framework. 
As shown in Figures~\ref{fig: cross-domain mmlu original} and \ref{fig: cross-domain mmlu sconu}, before conducting outliers detection and elimination within the test set, significant issues arise: the EMR exceeds the risk level (i.e., $\geq 0.1$) in testing datasets such as college chemistry, and many datasets report an APSS metric below 1, indicating that a substantial number of QA samples resulted in empty predictions. 
After filtering out test samples that significantly deviate from the calibration set, our foundational SConU framework achieves strict EMR control, with the APSS metrics of the test set nearly all at 1, highlighting that our method accurately identifies the correct answers.

On the more robust and challenging MMLU-Pro task with 10 options for each query, the minimum manageable risk level of each calibration set significantly increases, as presented in Table~\ref{table: mmlu-pro minimum risk level}. 
We set the risk level of $\alpha$ to 0.2 and present the results of the EMR metric utilizing the basic ConU framework in Figure~\ref{fig: cross-domain mmlu-pro original 32B}. 
The phenomenon of EMR surpassing the risk level is both frequent and severe. 
For example, when utilizing the Biology subset as the calibration to address queries from the Chemistry subject, the EMR metric is 0.56, significantly exceeding 0.2. 
Similarly, there are numerous QA samples where the prediction sets are empty, resulting in several subjects of the test sets having APSS scores below 1. 
Our SConU framework consistently maintains strict control over the EMR metric across all calibration-test set pairs, while also achieving higher prediction efficiency, as shown in Figure~\ref{fig: cross-domain mmlu-pro sconu 32B}. 
Furthermore, as demonstrated in Figure~\ref{fig: cross-domain mmlu-pro sconu-pro 32B}, SConU-Pro achieves lower EMR and APSS metrics across all calibration and test sets by assessing the reliability of the uncertainty scores of each calibration sample at a specific risk level.

\newpage
\input{Sections/Prompt_example}

%% file: Tables/MMLU.tex
\begin{table}[!h]
\centering
\caption{The number of samples employed for each subject from the MMLU dataset.}
\label{table: mmlu dataset}
\adjustbox{max width=\columnwidth}{
\begin{tabular}{lc}
\toprule
\textbf{Subjects} & \textbf{Number of Samples}\\
\midrule
computer security & 100\\
high school computer science & 100\\
college computer science & 100\\ 
machine learning & 112\\ 
formal logic & 126\\
\midrule
high school biology & 310\\ 
anatomy & 135\\
clinical knowledge & 265\\
college medicine & 173\\
professional medicine & 272\\
college chemistry & 100\\
\midrule
marketing & 234\\
public relations & 110\\
management & 103\\
business ethics & 100\\
professional accounting  & 282\\
\bottomrule
\end{tabular}
}
\end{table}

%% file: Tables/MMLU_Pro.tex
\begin{table}[!h]
\centering
\caption{The number of samples employed for each subject from the MMLU-Pro dataset. }
\label{table: mmlu-pro dataset}
\adjustbox{max width=\linewidth}{
\begin{tabular}{lc}
\toprule
\textbf{Subjects} & \textbf{Number of Samples}\\
\midrule
computer science\qquad\qquad\quad\quad & 410\\
math & 500\\
chemistry & 500\\
engineering & 500\\
law & 500\\
biology & 500\\
health & 500\\
physics & 500\\
business & 500\\
philosophy & 499\\
economics & 500\\
other & 500\\
psychology & 500\\
history & 381\\
\bottomrule
\end{tabular}
}
\end{table}

%% file: Tables/Hyperparameters.tex
\begin{table}[!t]
\centering
\adjustbox{max width=\linewidth}{
\begin{tabular}{ll}
\toprule
\textbf{Hyperparameter} & \textbf{Value}\\
\midrule
\texttt{do\_sample} & \texttt{True}\\
\texttt{num\_beams} & \texttt{1}\\
\texttt{top\_p} & \texttt{0.9}\\
\texttt{temperature\quad\quad\quad\quad} & \texttt{1.0}\\
\texttt{max\_length} & \texttt{input\_length + 1$/$36}\\
\bottomrule
\end{tabular}
}
\caption{Hyperparameters for the \texttt{generate} function. \texttt{input\_length} is the embedding length of the input prompt after being encoded by the tokenizer of the current language model.}
\label{table: hyperparameters}
\end{table}

%% file: Tables/mmlu_risk_level.tex
\begin{table*}[!t]
\centering
\caption{The minimum risk level manageable by each subject of the calibration set in the MMLU dataset utilizing the Qwen2.5-32B-Instruct model.}
\label{table: mmlu minimum risk level}
\adjustbox{max width=\linewidth}{
\begin{tabular}{lc|lc|lc}
\toprule
\textbf{Subjects (Computer Science)} & \textbf{$a_l$} & \textbf{Subjects (Medicine)} & \textbf{$a_l$} & \textbf{Subjects (Business)} & \textbf{$a_l$}\\
\midrule
computer security & 0 & high school biology & 0.007 & marketing & 0\\
high school computer science & 0 & anatomy & 0.009 & public relations & 0\\
college computer science & 0 & clinical knowledge & 0.004 & management & 0\\ 
machine learning & 0 & college medicine & 0.014 & business ethics & 0.011\\ 
formal logic & 0.019 & professional medicine & 0.008 & professional accounting  & 0.018\\
{} & {} & college chemistry & \textbf{0.051} & {} & {}\\
\bottomrule
\end{tabular}
}
\end{table*}

%% file: Tables/mmlu_pro_risk_level.tex
\begin{table}[!t]
\centering
\caption{The minimum risk level manageable by each subject of the calibration set in the MMLU-Pro dataset utilizing
the Qwen2.5-32B-Instruct model.}
\label{table: mmlu-pro minimum risk level}
\adjustbox{max width=\linewidth}{
\begin{tabular}{lc}
\toprule
\textbf{Subjects $\qquad\qquad\qquad\qquad\qquad\qquad\quad$} & \textbf{$a_l$}\\
\midrule
computer science\qquad\qquad\quad\quad & 0.075\\
math & 0.123\\
chemistry & 0.164\\
engineering & 0.126\\
law & 0.109\\
biology & 0.032\\
health & 0.047\\
physics & \textbf{0.161}\\
business & 0.112\\
philosophy & 0.045\\
economics & 0.030\\
other & 0.092\\
psychology & 0.015\\
history & 0.031\\
\bottomrule
\end{tabular}
}
\end{table}

%% file: Sections/Prompt_example.tex
\begin{figure*}[!t]
    \begin{tcolorbox}[title=MMLU]
$\#\#\#$ System:\\
Answer the following multiple-choice question by giving the most appropriate response. Answer should be one among [A, B, C, D].\\

$\#\#\#$ User:\\
What is penetration testing?\\
{A: A procedure for testing libraries or other program components for vulnerabilities; B: Whole-system testing for security flaws and bugs; C: A security-minded form of unit testing that applies early in the development process; D: All of the above}\\
$\#\#\#$ Assistant:\\
B\\

$\#\#\#$ User:\\
Suppose a user has an iPhone (running iOS) and downloads an app called Innocent from the Apple App Store and installs it. The user unlocks the phone and runs Innocent. Innocent exploits a bug in the iOS kernel which allows Innocent to redirect execution inside the kernel to code that Innocent controls. Now Innocent can execute any instructions it likes inside the iOS kernel. Innocent is not able to exploit any bugs in the phone’s secure enclave. Can Innocent read the user’s private information stored in the phone’s flash (e.g. Contacts and messages), or will the security measures described in the paper keep the data private? If Innocent is only able to see encrypted data, then the phone has successfully kept the data private. Circle the security features of the phone (if any) that will prevent Innocent from reading information from the flash on the phone.\\
{A: Secure boot chain; B: System software authorization; C: The secure enclave’s ephemeral key; D: None of the above}\\
$\#\#\#$ Assistant:\\
D\\

$\#\#\#$ User:\\
Why is it that anti-virus scanners would not have found an exploitation of Heartbleed?\\
{A: It's a vacuous question: Heartbleed only reads outside a buffer, so there is no possible exploit; B: Anti-virus scanners tend to look for viruses and other malicious; C: Heartbleed attacks the anti-virus scanner itself; D: Anti-virus scanners tend to look for viruses and other malicious code, but Heartbleed exploits steal secrets without injecting any code}\\
$\#\#\#$ Assistant:\\
D\\

$\#\#\#$ User:\\
Which of the following styles of fuzzer is more likely to explore paths covering every line of code in the following program?\\
{A: Generational; B: Blackbox; C: Whitebox; D: Mutation-based}\\
$\#\#\#$ Assistant:\\

\end{tcolorbox}
\caption{An example of the prompt in the MMLU task.}
\label{fig: mmlu prompt example}
\end{figure*}

\begin{figure*}[!t]
    \begin{tcolorbox}[title=MMLU-Pro]

$\#\#\#$ System:\\
Answer the following multiple-choice question by giving the most appropriate response. Answer should be one among [A, B, C, D, E, F, G, H, I, J].\\

$\#\#\#$ User:\\
In contrast to \underline{$\qquad$}, \underline{$\qquad$} aim to reward favourable behaviour by companies. The success of such campaigns have been heightened through the use of \underline{$\qquad$}, which allow campaigns to facilitate the company in achieving \underline{$\qquad$}.\\
{A: Boycotts, Buyalls, Blockchain technology, Increased Sales; B: Buycotts, Boycotts, Digital technology, Decreased Sales; C: Boycotts, Buycotts, Digital technology, Decreased Sales; D: Buycotts, Boycotts, Blockchain technology, Charitable donations; E: Boycotts, Buyalls, Blockchain technology, Charitable donations; F: Boycotts, Buycotts, Digital technology, Increased Sales; G: Buycotts, Boycotts, Digital technology, Increased Sales; H: Boycotts, Buycotts, Physical technology, Increased Sales; I: Buycotts, Buyalls, Blockchain technology, Charitable donations; J: Boycotts, Buycotts, Blockchain technology, Decreased Sales}\\
$\#\#\#$ Assistant:\\
F\\

$\#\#\#$ User:\\
\underline{$\qquad$} is the direct attempt to formally or informally manage ethical issues or problems, through specific policies, practices and programmes.\\
{A: Operational management; B: Corporate governance; C: Environmental management; D: Business ethics management; E: Sustainability; F: Stakeholder management; G: Social marketing; H: Human resource management; I: N/A; J: N/A}\\
$\#\#\#$ Assistant:\\
D\\

$\#\#\#$ User:\\
How can organisational structures that are characterised by democratic and inclusive styles of management be described?\\
{A: Flat; B: Bureaucratic; C: Autocratic; D: Hierarchical; E: Functional; F: Decentralized; G: Matrix; H: Network; I: Divisional; J: Centralized}\\
$\#\#\#$ Assistant:\\
A\\

$\#\#\#$ User:\\
Typical advertising regulatory bodies suggest, for example that adverts must not: encourage \underline{$\qquad$}, cause unnecessary \underline{$\qquad$} or \underline{$\qquad$}, and must not cause \underline{$\qquad$} offence.\\
{A: Safe practices, Fear, Jealousy, Trivial; B: Unsafe practices, Distress, Joy, Trivial; C: Safe practices, Wants, Jealousy, Trivial; D: Safe practices, Distress, Fear, Trivial; E: Unsafe practices, Wants, Jealousy, Serious; F: Safe practices, Distress, Jealousy, Serious; G: Safe practices, Wants, Fear, Serious; H: Unsafe practices, Wants, Fear, Trivial; I: Unsafe practices, Distress, Fear, Serious}\\
$\#\#\#$ Assistant:\\

\end{tcolorbox}
\caption{An example of the prompt in the MMLU-Pro task. Note that the current question has 9 options.}
\label{fig: mmlu-pro prompt example}
\end{figure*}

\begin{figure*}[!t]
    \begin{tcolorbox}[title=MedMCQA]

$\#\#\#$ System:\\
Answer the following multiple-choice question by giving the most appropriate response. Answer should be one among [A, B, C, D].\\

$\#\#\#$ User:\\
Kamlesh, a 2 year old girl, has Down's syndrome. Her karyotype is 21$/$21 translocation. What is the risk ofrecurrence in subsequent pregnancies if the father is a balanced translocation carrier :\\
{A: 100$\%$; B: 50$\%$; C: 25$\%$; D: 0$\%$}\\
$\#\#\#$ Assistant:\\
A\\

$\#\#\#$ User:\\
Not a part of ethmoid bone is\\
{A: Inferior turbinate; B: Agar nasi cells; C: Uncinate process; D: Crista galli}\\
$\#\#\#$ Assistant:\\
A\\

$\#\#\#$ User:\\
Haddon matrix is related to:\\
{A: Injury prevention; B: Communicable diseases; C: Maternal and child moality; D: Hypeensive disorders}\\
$\#\#\#$ Assistant:\\
B\\

$\#\#\#$ User:\\
Which of the following is not true for myelinated nerve fibers:\\
{A: Impulse through myelinated fibers is slower than non-myelinated fibers; B: Membrane currents are generated at nodes of Ranvier; C: Saltatory conduction of impulses is seen; D: Local anesthesia is effective only when the nerve is not covered by myelin sheath}\\
$\#\#\#$ Assistant:\\

\end{tcolorbox}
\caption{An example of the prompt in the MedMCQA task.}
\label{fig: MedMCQA prompt example}
\end{figure*}

\begin{figure*}[!t]
    \begin{tcolorbox}[title=TriviaQA]

$\#\#\#$ System:\\
This is a bot that correctly answers questions.\\

$\#\#\#$ User:\\
In 1968, who did radical feminist Valerie Solanas shoot and wound as he entered his New York studio?\\
$\#\#\#$ Assistant:\\
Andy Warhol\\

$\#\#\#$ User:\\
What lake can be found on the border of Vermont and New York?\\
$\#\#\#$ Assistant:\\
Lake Champlain\\

$\#\#\#$ User:\\
Which competition was won by Nadiya Hussain in 2015?\\
$\#\#\#$ Assistant:\\
The Great British Bake-Off\\

$\#\#\#$ User:\\
Who was the man behind The Chipmunks?\\
$\#\#\#$ Assistant:\\

\end{tcolorbox}
\caption{An example of the prompt in the TriviaQA task.}
\label{fig: TriviaQA prompt example}
\end{figure*}

\begin{figure*}[!t]
    \begin{tcolorbox}[title=CoQA]

$\#\#\#$ System:
This is a bot that correctly answers questions.\\
Once upon a time, in a barn near a farm house, there lived a little white kitten named Cotton. Cotton lived high up in a nice warm place above the barn where all of the farmer's horses slept. But Cotton wasn't alone in her little home above the barn, oh no. She shared her hay bed with her mommy and 5 other sisters. All of her sisters were cute and fluffy, like Cotton. But she was the only white one in the bunch. The rest of her sisters were all orange with beautiful white tiger stripes like Cotton's mommy. Being different made Cotton quite sad. She often wished she looked like the rest of her family. So one day, when Cotton found a can of the old farmer's orange paint, she used it to paint herself like them. When her mommy and sisters found her they started laughing. ``What are you doing, Cotton?!'' ``I only wanted to be more like you''. Cotton's mommy rubbed her face on Cotton's and said ``Oh Cotton, but your fur is so pretty and special, like you. We would never want you to be any other way''. And with that, Cotton's mommy picked her up and dropped her into a big bucket of water. When Cotton came out she was herself again. Her sisters licked her face until Cotton's fur was all all dry. ``Don't ever do that again, Cotton!'' they all cried. ``Next time you might mess up that pretty white fur of yours and we wouldn't want that!'' Then Cotton thought, ``I change my mind. I like being special''.\\

$\#\#\#$ User:\\
What color was Cotton?\\
$\#\#\#$ Assistant:\\
white\\

$\#\#\#$ User:\\
Where did she live?\\
$\#\#\#$ Assistant:\\
in a barn\\

$\#\#\#$ User:\\
Did she live alone?\\
$\#\#\#$ Assistant:\\
no\\

$\#\#\#$ User:\\
Who did she live with?\\
$\#\#\#$ Assistant:\\

\end{tcolorbox}
\caption{An example of the prompt in the CoQA task.}
\label{fig: CoQA prompt example}
\end{figure*}